\documentclass[pdflatex,sn-mathphys]{sn-jnl}

\jyear{2021}%

\theoremstyle{thmstyleone}%
\theoremstyle{thmstyletwo}%

\theoremstyle{thmstylethree}%

\begin{document}

\title[EV-MgNet for Operator Learning in Numerical PDEs]{An Enhanced V-cycle MgNet Model for Operator Learning in Numerical Partial Differential Equations}


\author[1]{\fnm{Jianqing} \sur{Zhu}}\email{jqzhu@emails.bjut.edu.cn}

\author[2]{\fnm{Juncai} \sur{He}}\email{juncai.he@kaust.edu.sa}
\equalcont{This author contributed equally to this work.}

\author*[1]{\fnm{Qiumei} \sur{Huang}}\email{qmhuang@bjut.edu.cn}

\affil*[1]{\orgdiv{Faculty of Sciences}, \orgname{Beijing University of Technology}, \orgaddress{\street{Chaoyang}, \city{Beijing}, \postcode{100124}, \state{Beijing}, \country{People’s Republic of China}}}

\affil[2]{\orgdiv{Computer, Electrical and Mathematical Science and Engineering Division}, \orgname{King Abdullah University of Science and Technology (KAUST)}, \orgaddress{\city{Thuwal}, \postcode{23955-6900}, \country{Saudi Arabia}}}


\abstract{This study used a multigrid-based convolutional neural network architecture known as MgNet in operator learning to solve numerical partial differential equations (PDEs). 
Given the property of smoothing iterations in multigrid methods where low-frequency errors decay slowly, we introduced a low-frequency correction structure for residuals to enhance the standard V-cycle MgNet. 
The enhanced MgNet model can capture the low-frequency features of solutions considerably better than the standard V-cycle MgNet. The numerical results obtained using some standard operator learning tasks are better than those obtained using many state-of-the-art methods, demonstrating the efficiency of our model.
Moreover, numerically, our new model is more robust in case of low- and high-resolution data during training and testing, respectively.}




\maketitle
\section{Introduction}
Partial differential equations (PDEs) are critical in physical and engineering problems~\cite{courant2008methods}.
Thus, solving PDEs quickly and accurately is critical in practice. 
Currently, there are many mature numerical algorithms for solving PDEs, such as finite element methods~\cite{brenner2008mathematical} and finite difference methods~\cite{leveque2007finite}. 
Typically, these methods discretize continuous PDEs, then, discrete linear or nonlinear systems are solved using iterative or direct methods.
However, traditional methods have the drawback of having to repeat the discretization and solving process if the parameters or initial conditions of PDEs change. 
In addition, designing and implementing fast solvers for different types of discretized PDE systems is difficult. 
Thus, approximately constructing the operator from the parameters or conditions of PDEs with respect to the solutions, which is also known as  operator learning, has attracted increasing interest in the case of numerical PDEs~\cite{lu2019deeponet,li2020fourier,nelsen2021random,gupta2021multiwavelet}. 

In recent years, methods based on deep learning models~\cite{lecun2015deep} have achieved great success surpassing humans in the fields of computer vision~\cite{he2016deep}, natural language processing \cite{vaswani2017attention}, and reinforcement learning~\cite{silver2016mastering}. 
Researchers have proposed applying deep learning methods to science and engineering problems based on these inspirational applications of deep learning in artificial intelligence. 
In this study, we focus on operator learning in numerical PDEs with neural network architecture.  

The studies \cite{guo2016convolutional,zhu2018bayesian,adler2017solving,bhatnagar2019prediction,khoo2021solving,chen2022meta} in this field mainly use convolutional neural networks (CNNs) to achieve the mapping between two discretized function spaces. The main disadvantage of these methods is that many CNN-based models depend on the data's discrete scale and may only work at a specific resolution~\cite{kovachki2021neural}.
To solve this problem, researchers have proposed many new network structures~\cite{lu2019deeponet,bhattacharya2020model,li2020fourier,nelsen2021random,patel2021physics,anandkumar2020neural,cao2021choose,gupta2021multiwavelet,liu2022ht} and demonstrated their results using some more well-prepared data sets for operator learning in case of numerical PDEs.
In~\cite{lu2019deeponet,goswami2021physics,di2021deeponet}, a neural network architecture known as DeepONet, was proposed to construct the mapping from a discrete input to a continuous function which is interpreted as a continuous approximation to the real solution.
In~\cite{li2020fourier}, the authors proposed Fourier neural operator (FNO) which parametrizes the integral kernel directly in Fourier space and uses only low-frequency information. Subsequently,~\cite{gupta2021multiwavelet} proposed a multiwavelet-based neural operator learning scheme, which compresses the kernel using wavelet bases. 
Based on the massive success of Transformer~\cite{vaswani2017attention} in deep learning, Transformer-based operator learning architecture was also introduced in~\cite{cao2021choose,liu2022ht}.

In this study, we develop an efficient operator learning model based on the MgNet structure, which has been enhanced via a new basic smoothing iteration (smoother).
MgNet~\cite{he2019mgnet,he2021interpretive} was first proposed as a uniform framework for CNNs for image classification~\cite{he2016deep,he2016identity} and multigrid methods for solving numerical PDEs~\cite{xu1992iterative,xu2002method,hackbusch2013multi,xu2017algebraic}.
Motivated by the critical feature of smoothing iterations in multigrid, in which high frequencies of error decay first, we suggest enhancing the basic smoothing iterations (basic blocks) in MgNet with a low-frequency correction part.
To appropriately deal with low frequencies during MgNet iterations, we used the Fourier transform to select low frequencies and multiply a learnable fully connected layer (frequency-wise) to reduce the error in the low-frequency domain. 
This process is similar to applying an FNO to residuals in MgNet.
Therefore, we propose a new MgNet-based operator learning model improved with a trainable low-frequency residual correction structure. Numerically, the enhanced MgNet model can reduce the low-frequency error considerably better than the standard MgNet model. 
In terms of general benchmarks in operator learning for numerical PDEs, our new model achieves state-of-the-art results in the one-dimensional (1D) Burgers' equation, the 1D nonlinear Korteweg-de Vries equation, the two-dimensional (2D) Darcy flow equation, and the 2D Navier-Stokes equations. Furthermore, under learning at lower resolutions, our model can generalize to high resolutions, which is difficult for popular methods such as FNO and MWT~\cite{li2020fourier,kovachki2021universal,lanthaler2022error}.

The paper is organized as follows. In section~\ref{sec:ol}, we introduce the neural operator learning methods using V-cycle MgNet and FNO as examples. In section~\ref{sec:model}, we propose the main model developed by enhancing MgNet with low-frequency correction architecture. In section~\ref{sec:numerics}, we demonstrate the efficiency of our proposed model regarding accuracy and robustness on different PDEs data sets and resolutions. Finally, in section~\ref{sec:conclusion}, we add some concluding remarks.

\section{Neural operators for operator learning}\label{sec:ol}
In this section, we first define the operator learning problem in the case of numerically solving PDEs. 
Then, we introduce the V-cycle MgNet as an example 
of a typical CNN-based operator learning architecture. Finally, we present a special neural network known as the Fourier neural operator (FNO), which is closely related to our low-frequency correction structure.
\subsection{Operator learning for numerical PDEs}
In this study, we mainly consider two types of numerical PDE problems: time-dependent and stationary PDEs. For time-dependent PDEs, we consider the following initial-boundary value problem:
\begin{equation}
	\begin{cases}
		&\mathcal L(u, u_t, \nabla u, \nabla^2 u) = 0, \quad x\in\Omega, t>0, \\
		&u(x,0) = \mu(x), \quad x\in\Omega.
	\end{cases}
\end{equation}
with the periodic boundary condition on $\Omega = (0,1)^d$ where $d=1,2$. For stationary PDEs, we consider a boundary value problem:
\begin{equation}
	\begin{cases}
		&\mathcal L(u, \nabla u, \nabla^2 u; a) = 0, \quad x\in\Omega, \\
		&u(x) = 0, \quad x\in\partial\Omega.
	\end{cases}
\end{equation}
where $\Omega=(0,1)^2$ and $a = a(x)$ is the parameter function or the source term in the PDE.
The operator learning question in a continuous level is to approximate an operator from the initial condition $\mu(x)$ to the solution at a fixed time $T$, i.e., $\mu(x) \mapsto u(x,T)$ or an operator from the parameter function $a(x)$ to the solution $u(x)$. In the discrete level, we assume that there is a grid (typically a uniform mesh) $\mathcal T_h$ on $\Omega$ such that $\mu$, $a$, $u(T)$, and $u$ represent tensors formed by all the discretized values at the grid points of $\mu(x)$, $a(x)$, $u(x,T)$, and $u(x)$ respectively. For example, one may have $\mu, u(T) \in \mathbb R^{d_x}$ for 1D questions and $a, u \in \mathbb R^{d_x\times d_y}$ for 2D questions. Thus, the operator learning task can be formulated as learning (approximating) a map $\mathcal M$ from a tensor to another tensor with the same spatial dimensions.
By the abuse of a notation, $a$ is used to denote the input tenor and $u$ is used to denote the output tensor. Here, $a$ can denote initial conditions, parameter functions, source terms, etc. Similarly, $u$ can denote the solution at a certain time or the solution to a stationary problem.
For simplicity, we assume that the data set is $\left\{ a_i, u_i \right\}_{i=1}^N$. The inputs $a_i$ are sampled i.i.d from a distribution $\mathcal D$ for all $i=1:N$, i.e. $a_i \sim \mathcal D $. $u_i$ denotes solutions that are obtained via specific classical numerical methods.
The operator learning task can be formulated as the following optimization problem
\begin{equation}\label{eq:loss}
	\min_{\theta} \frac{1}{N}\sum_{i=1}^N \frac{\|\mathcal M(a_i;\theta) - u_i\|^2}{\|u_i\|^2},
\end{equation}
where $\theta$ denotes the parameters of the model to construct the operator $\mathcal M(\cdot)$ and $\| \cdot \|$ defines the $l^2$ norm of the tensor.
More details about the data set and model parameters are presented in section \ref{sec:numerics}.

\subsection{V-cycle MgNet for operator learning}
The multigrid (MG) method is one of the most efficient methods for solving differential equations using a discretization hierarchy.
This method provides an efficient approach for approximating the inverse of a discretized elliptic differential operator.
Thus, applying the MG method to the operator learning tasks is natural.
Recently, MgNet~\cite{he2019mgnet} was proposed as a uniform framework for MG methods and CNNs.
In addition, \cite{he2022approximation} demonstrated that the CNN model with MgNet architecture could have a universal approximation power.
These two observations motivate us to use the MgNet for operator learning tasks. 
Consequently, we used the V-cycle MgNet as the basic architecture for learning CNN-based neural operators. 

Generally, the neural operator method \cite{anandkumar2020neural} is an iterative architecture that uses neural networks to build a tensor-to-tensor map $\mathcal M$ from the discretized input $a$ to output $u$
as $a \rightarrow u^0 \rightarrow u^1 \rightarrow \dots \rightarrow u^L \rightarrow u$. 
For simplicity, consider a standard V-cycle version of MgNet (V-MgNet) applied to a 2D operator learning problem, where the input $a$ is a discretized 2D function with $d\times d$ sampling points on the uniform mesh and one channel, i.e., $a \in \mathbb R^{d\times d \times 1}$, and the output $u$ is the discretized solution function on the same mesh typically with only one channel, i.e., $u \in \mathbb R^{d\times d \times 1}$. 
Although there is only one channel in the input $a$ and output $u$, more channels are essential for the success of CNNs. In~\cite{he2022approximation}, the authors provided a quantitative relationship between the number of channels and the approximation power of CNNs.
Thus, we need to introduce channels in V-MgNet. 
Given the numerical results in MgNet~\cite{he2021interpretive,wang2022cnns} in image classification, we set a fixed channel number in different grids and layers, which can accelerate the process of fine-tuning the hyper-parameters. Here, we present the V-cycle MgNet (V-MgNet) algorithm.
\begin{algorithm}[H]
	\caption{$u=\text{V-MgNet}(a;J,\nu_\ell,c,C)$}
	\label{alg:mgnet}
	\begin{algorithmic}[1]
			\State {\bf Input}: discretized parameter function (for example,$a \in \mathbb R^{d\times d \times 1}$), number of grids $J$, number of smoothing iterations $\nu_\ell$ for $\ell=1:J$, number of channels $c$ on each grid, and number of channels $C$.
			\State {\bf Initialization}: $f^1 = \sigma \circ K^0\ast a \in \mathbb{R}^{d_1\times d_1 \times c}$, $u^{1,0}=0$, and discretized spatial size $d_\ell = \frac{d}{2^{\ell-1}}$ for $\ell=1:J$.
			\For{$\ell = 1:J$}
			\State Feature extraction (smoothing):
			\For{$i = 1:\nu_\ell$}
			\State 
			\begin{equation}\label{eq:mgiteration}
				u^{\ell,i}=u^{\ell,i-1}+ \sigma\circ B^{\ell,i} \ast \sigma\left(f^{\ell}-A^{\ell}\ast u^{\ell,i-1}\right) \in \mathbb{R}^{d_\ell\times d_\ell \times c}. 
			\end{equation}
			\EndFor
			
			\State Note:
			$
			u^\ell= u^{\ell,\nu_\ell}
			$
			\If{$\ell<J$}
			\State Interpolation and restriction:
			$$
			\begin{aligned}
				u^{\ell+1,0}&= \Pi^{\ell+1}_{\ell}\ast_2u^{\ell} \in \mathbb{R}^{d_{\ell+1}\times d_{\ell+1} \times c}, \\ f^{\ell+1}&= R_{\ell}^{\ell+1}\ast_2\left(f^{\ell}-A^{\ell} u^{\ell}\right)+ A^{\ell+1}\ast u^{\ell+1,0}  \in \mathbb{R}^{d_{\ell+1}\times d_{\ell+1} \times c}.
			\end{aligned}
			$$
			\EndIf
			\EndFor
			
			\For{$\ell = J-1:1$}
			\State Prolongation:
			$$u^{\ell, 0} = u^{\ell}+ P_{\ell+1}^{\ell}\ast^2 (u^{\ell+1}-u^{\ell+1,0}) \in \mathbb{R}^{d_{\ell}\times d_{\ell} \times c} $$
			\For{$i = 1:\nu_\ell$}
			\State Feature extraction (smoothing):
			\begin{equation*}
				u^{\ell,i} = u^{\ell,i-1} + \sigma\circ B^{\ell,i} \ast \sigma \left(f^{\ell}-A^{\ell}\ast u^{\ell,i-1}\right) \in \mathbb{R}^{d_{\ell}\times d_{\ell} \times c}.
			\end{equation*}
			\EndFor
			
			\EndFor
			\State {\bf Output:}
			$$u = K^{2} \ast \sigma \circ K^1 \ast u^{1,\nu_1} \in \mathbb{R}^{d\times d \times 1 }.$$
		\end{algorithmic}
	\end{algorithm}
	
	We explained the notations in the above algorithm using an example of 2D PDEs (i.e. $a,u\in \mathbb{R}^{d\times d\times 1}$). For 1D PDEs, we only revise the filter structure from 2D to 1D by keeping the filter size accordingly.
	As in CNNs, $K^{i}$ for $i=0:2$ denote the classical $1\times1$ convolution operations with multichannel and stride 1. $K^0$ raises the number of channels from $1$ to $c$, $K^1$ maps the number of channels from $c$ to $C$, and $K^2$ maps the number of channels from $C$ back to $1$.
	$A^\ell\ast$ and $B^\ell\ast$ denote the classical $3\times3$ convolution operations with multichannel and stride 1, which correspond to the discretized system and the smoother in the MG methods for numerical PDEs.
	Here, $\Pi^{\ell+1}_{\ell}\ast_2$ and $R_{\ell}^{\ell+1}\ast_2$ denote the $3\times3$ convolution operations with multichannel and stride 2 which are known as pooling operators in CNNs but play the same roles of interpolation and restriction in the MG methods. 
	Correspondingly, we denote $P_{\ell+1}^{\ell}\ast^2$ as the $3\times 3$ deconvolution operation with multichannel and stride 2 which defines the prolongation operator in MG. 
	Theoretically, the interpolation and restriction operators in standard geometric MG methods~\cite{xu1992iterative,hackbusch2013multi,xu2017algebraic} can be represented as the convolution operators with stride 2, as shown in~\cite{he2019mgnet}. 
	In addition, $\sigma(\cdot)$ or $\sigma\circ$ denotes the element-wise application of the activation function $\sigma(x)$.
	
	Algorithm \ref{alg:mgnet} provides a uniform framework for MG methods and CNNs.
	When $\sigma(\cdot)$ is ignored, Algorithm \ref{alg:mgnet} is a standard MG method, while when $\sigma(\cdot)$ is added, Algorithm \ref{alg:mgnet} is MgNet as a CNN. These properties illustrate the close connection between the MG method and CNNs.

	\subsection{Fourier neural operator}
	Recently, a special type of neural operator based on Fourier transformation was proposed to capture the low frequencies of the solution $u$ using the information obtained from the low frequencies of the input $a$. For simplicity, consider the 1D operator learning problem with one channel for input and output, i.e., $a, u \in \mathbb R^{d\times 1}$. The FNO \cite{li2020fourier} applies the update rule $u^{\ell-1}\rightarrow u^{\ell}$ as
	\begin{equation}\label{eq:fno}
		u^{\ell} =\sigma\left(K^\ell \ast u^{\ell-1}+\mathcal{K}^\ell u^{\ell-1}\right),
	\end{equation}
	where $\sigma$ is the activation function, $K^\ell$ denotes a convolution with a multichannel and filter size $1$, and $\mathcal{K}^\ell$ represents the trainable Fourier integral operator defined as
	\begin{equation}
		\mathcal{K}^\ell u^{\ell-1}=\mathcal{F}^{-1} \mathcal W^\ell\mathcal{F} u^{\ell-1}.
	\end{equation}
	In this equation, $\mathcal{F}$ denotes the Fourier transform and $\mathcal{F}^{-1}$ represents the inverse transform and $\mathcal W^\ell$ is a linear operator defined as a combination of truncation and non-sharing convolution with a multichannel and a filter size $1$. Specifically, if $u^{\ell-1} \in \mathbb R^{d \times c}$, we have $\mathcal F \in \mathbb C^{d\times d}$ and $\widehat{u}^{\ell-1} = \mathcal F u^{\ell-1} \in \mathbb C^{d\times c}$ is the result of applying $\mathcal F$ to each channel of $u^{\ell-1}$.
	Here, $c$ denotes the number of channels for $u^\ell$, which is also known as the number of features in FNO~\cite{li2020fourier} and other relevant literature. 
	Correspondingly, $\mathcal F^{-1}$ defines a channel-wise Fourier inverse transform.
	Then, $\mathcal W^\ell: \mathbb C^{d\times c} \mapsto \mathbb C^{d\times c}$, the most important operator in FNO~\cite{li2020fourier}, is defined as 
	\begin{equation}\label{eq:fno_w}
		\left[\mathcal W^{\ell} \widehat{u}^{\ell-1}\right]_{k,i}=
		\begin{cases}
			\sum_{j=1}^{c}W^{\ell}_{k,i,j}\widehat{u}^{\ell-1}_{k,j}& k\leq k_{\rm max}\\
			0& k>k_{\rm max}
		\end{cases},
	\end{equation}
	where $W^\ell_{k,i,j} \in \mathbb R$ for $k=1:d$ and $i,j=1:c$ are the trainable parameters of the Fourier integral operator $\mathcal K^\ell$. That is, only the smallest $k_{\rm max}$ Fourier modes are used. Specifically, $\mathcal K^\ell$ constructs only the smallest $k_{\rm max}$ Fourier modes of $u^{\ell}$ using only the information obtained from the $k_{\rm max}$ lowest frequencies of $u^{\ell-1}$.
	
	We present the FNO~\cite{li2020fourier} algorithm as follows:
	\begin{algorithm}[H]
		\caption{$u=\text{FNO}(a, k_{\rm max}, L, c, C)$}
		\label{alg:fno}
		\begin{algorithmic}[1]
			
			\State {\bf Input:} discretized parameter function $a$, the number of lowest Fourier modes $k_{\rm max}$, number of layers $L$, number of channels $c$ in hidden layers, number of channels $C$ in the last output layer.
			\State {\bf Initialization:}  $u^0 =K^0\ast a$
			\For{$\ell = 1:L$}
			\State 
			\begin{equation*}
				u^{\ell}=\sigma\left(K^\ell \ast u^{\ell-1}+ \mathcal K^\ell u^{\ell-1}  \right), \quad \mathcal K^\ell = \mathcal{F}^{-1} \mathcal W^\ell \mathcal{F}.
			\end{equation*}
			\EndFor
			\State{\bf Output:}
			$$u = K^{L+2} \ast \sigma \circ K^{L+1}\ast {u}^{L}.$$
			
		\end{algorithmic}
	\end{algorithm}
	
	In the FNO~\cite{li2020fourier} code, $K^{\ell}$ for $1:\ell$ is considered as the classic convolution with a multichannel, a filter size of $1$, and a stride of 1. Moreover, $K^{0}$ raises the channel from $1$ to $c$, $K^{\ell}$ for $\ell=1:L$ maintains the channel number as $c$, $K^{L+1}$ changes the channel number from $c$ to $C$, and finally $K^{L+2}$ reduces the channel number from $C$ back to $1$ to fit the channel number of $u$. The choices of $c$ and $C$ are determined using the spatial dimensions of $a$ and $u$.
	Similar to the V-MgNet, we demonstrate the notation of FNO by using 1D data. For 2D problems, the definitions can be correspondingly obtained.
	
	The difference between FNO and the V-MgNet is that FNO has only one spatial resolution along all the layers (i.e., there are no pooling operations in FNO), whereas the multiscale spatial resolutions play a critical role in V-MgNet. In addition, FNO has a fixed number of the lowest frequencies. In our model, we used the MG structure and various numbers of the smallest Fourier modes.
	
\section{Enhanced V-MgNet with low-frequency correction}\label{sec:model}
\begin{figure}[H]
	\centerline{V-MgNet}
	\centerline{\includegraphics[width=.6\textheight]{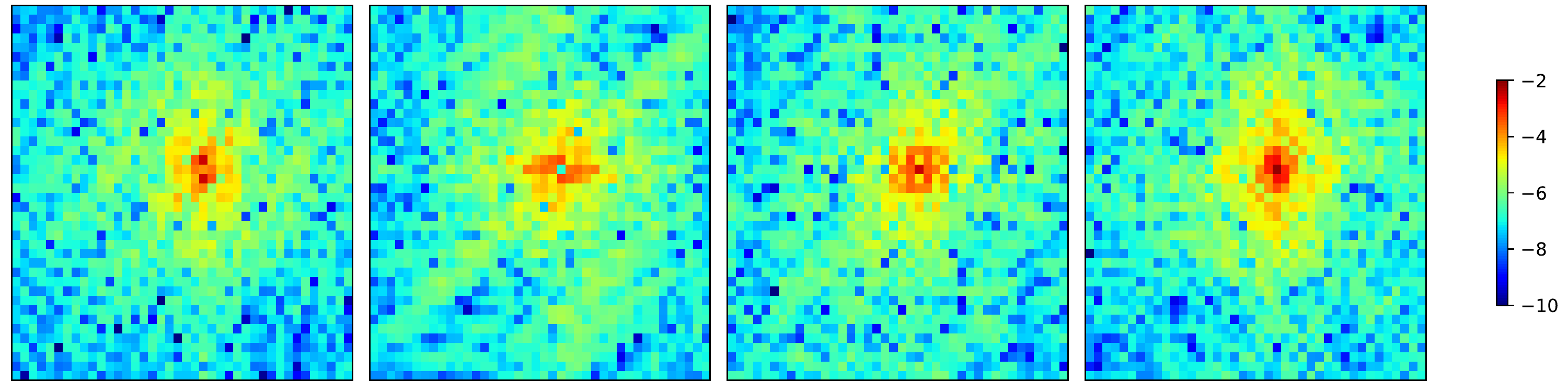}}
	\centerline{EV-MgNet}
	\centerline{\includegraphics[width=.6\textheight]{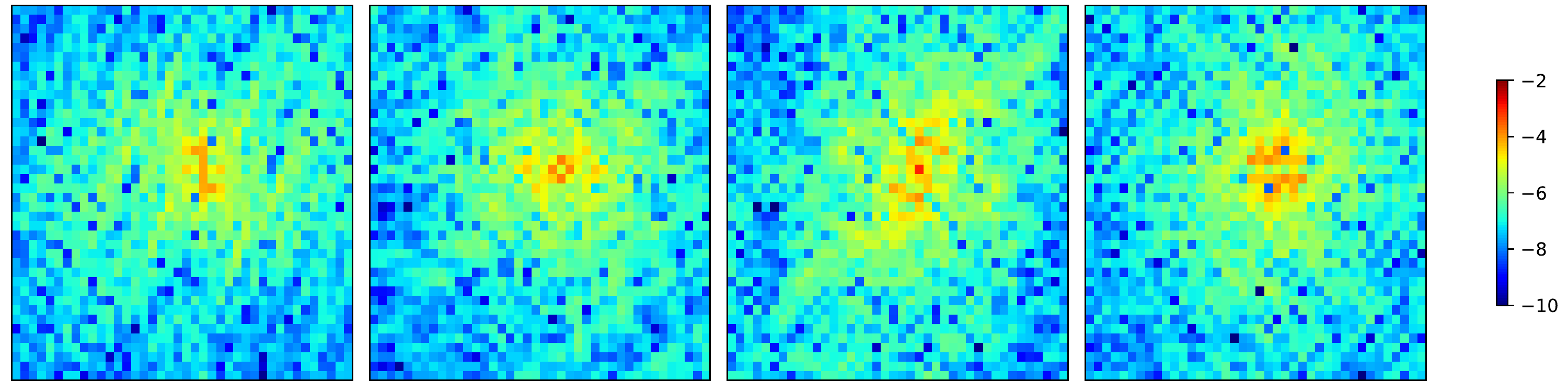}}
	\caption{Four random examples of the error $e_i = \mathcal M(a_i;\theta) - u_i$ in case of relatively low frequencies using the 2D Darcy flow data set (more details can be found in Section~\ref{sec:numerics}), where $\mathcal M(\cdot, \theta)$ is the V-MgNet in Algorithm~\ref{alg:mgnet} (above) or the EV-MgNet in presented Algorithm~\ref{alg:mgfno} (below).}
	\label{fig:Fourier-error}
\end{figure}

In this section, we propose an operator learning architecture by enhancing the
V-cycle MgNet with a low-frequency correction structure. Using this specially designed structure in V-MgNet, we obtain a more accurate iterative block, with a considerably improved performance than the existing models.
The frequency arguments in iterative methods for discretized PDEs systems~\cite{trottenberg2000multigrid,xu2017algebraic} motivated us to perform this modification.
For simplicity, we first ignore the activation function $\sigma$ in V-MgNet. Then, the solution can be obtained using the following residual correction scheme with an algebraic system $Au=f$:
\begin{equation}\label{eq:residual}
	u^{i+1} = u^{i}+B(f-Au^{i}),
\end{equation}
where $B \approx A^{-1}$ is know as the smoother of the iterative method. 
As shown in \cite{he2019mgnet}, $A$ and $B$ can be written as some convolutional kernels if $A$ is a discretized elliptic partial differential operator on uniform mesh and $B$ is the corresponding smoother, such as the one-step Jacobi iteration.
Using the notation of convolution, \eqref{eq:residual} can be written as 
\begin{equation}\label{eq:ite}
	u^{i+1} = u^{i}+B\ast (f-A\ast u^{i-1}).
\end{equation}
A fundamental property for the above-mentioned iterative scheme \eqref{eq:ite} is that the high frequencies of the error $e^i = u - u^i$ decay rapidly during the iteration process (as $i$ increases) while the low frequencies decay slowly. 
As shown in Figure~\ref{fig:Fourier-error}, even for the (non-linear) V-MgNet in Algorithm~\ref{alg:mgnet} for operator learning, the low frequencies of the error $e=$V-MgNet($a$)$-u$ are dominated.

Thus, we propose to add a low-frequency enhancement part into \eqref{eq:ite}, which can be understood as an FNO-type operator for the residual. 
In addition, we introduce the following iterative scheme:
\begin{equation}\label{eq:mgfno_it}
	u^{i+1} = u^{i}+B\ast (f-Au^{i}) + \mathcal{F}^{-1}\mathcal{W}^i\mathcal{F} (f-Au^{i}),
\end{equation}
where $\mathcal{W}^i$ follows the same definition in~\eqref{eq:fno_w} which operates on low frequencies only. 
We obtained the following basic iterative scheme for our new model based on the $\ell-$th level for $i-$th iteration by introducing the activation function into the new iterative structure in~\eqref{eq:mgfno_it}.
\begin{equation}\label{eq:mgfno_it_a}
	u^{\ell,i} = u^{\ell,i-1}+\sigma \circ B^{\ell,i}\ast \sigma (r^{i-1}) + \sigma \circ \mathcal{F}^{-1}\mathcal{W}^{\ell,i}\mathcal{F} (r^{\ell,i-1}),
\end{equation}
where $r^{\ell,i-1} = f^{\ell}-A^{\ell}\ast u^{\ell,i-1}$ and $\sigma(\cdot)$ denotes the activation function. 
We present our enhanced V-MgNet (EV-MgNet) by combining all the previous descriptions as follows.  

\begin{algorithm}[H]
	\caption{$u=\text{EV-MgNet operator}(a; J,\nu_\ell,k^\ell_{\rm max},c,C)$}
	\label{alg:mgfno}
	\begin{algorithmic}[1]
		\State {\bf Input:} discretized parameter function $a$, the number of grids $J$, number of smoothing iterations $\nu_\ell$, number of lowest Fourier modes $k_{\rm max}$ for $\ell=1:J$, number of channels $c$ on each grid, and number of channels $C$.
		\State {\bf Initialization:} discretized spatial size $d_\ell = \frac{d}{2^{\ell-1}}$ for $\ell=1:J$, $f^1 = \sigma \circ K^0 \ast f$, $u^{1,0}=0$
		\For{$\ell = 1:J$}
		\For{$i = 1:\nu_\ell$}
		\State Feature extraction (smoothing):
		\begin{equation*}
			u^{\ell,i} = u^{\ell,i-1} + \sigma \circ B^{\ell,i} \ast \sigma\left(r^{\ell,i-1}\right) +\sigma \circ \mathcal{F}^{-1}\mathcal{W}^{\ell,i}\mathcal{F} \left(r^{\ell,i-1}\right), 
		\end{equation*}
		\quad \quad \quad where $r^{\ell,i-1}=f^{\ell}-A^{\ell}\ast u^{\ell,i-1}$.
		\EndFor
		\State Note:
		$
		u^\ell= u^{\ell,\nu_\ell}
		$
		\If{$\ell<J$}
		\State Interpolation and restriction:
		\begin{equation*}
			\begin{aligned}
				u^{\ell+1,0}&= \Pi^{\ell+1}_{\ell}\ast_2 u^{\ell}, \\ f^{\ell+1}&= R_{\ell}^{\ell+1} \ast_2 \left(f^{\ell}-A^{\ell} \ast u^{\ell}\right) + A^{\ell+1}\ast u^{\ell+1,0}.
			\end{aligned}
		\end{equation*}
		\EndIf
		\EndFor
		
		\For{$\ell = J-1:1$}
		
		\State Prolongation:
		\begin{equation*}
			u^{\ell,0}=u^{\ell}+ P_{\ell+1}^{\ell} \ast^2(u^{\ell+1}-u^{\ell+1,0}).
		\end{equation*}
		\For{$i = 1:\nu_\ell$}
		\State Feature extraction (smoothing):
		\begin{equation*}
			u^{\ell,i} = u^{\ell,i-1} + \sigma \circ B^{\ell,i} \ast \sigma\left(r^{\ell,i-1}\right) +\sigma \circ \mathcal{F}^{-1}\mathcal{W}^{\ell,i}\mathcal{F}\left(r^{\ell,i-1}\right),
		\end{equation*}
		\quad \quad \quad where $r^{\ell,i-1}=f^{\ell}-A^{\ell}\ast u^{\ell,i-1}$.
		\EndFor
		\EndFor
		
		\State {\bf Output:}
		$$
		u = K^{2} \ast \sigma \circ K^1 \ast u^{1,\nu_1}.
		$$
	\end{algorithmic}
\end{algorithm}

Generally, the notations presented in the above-mentioned algorithm have the same definitions as those for the V-MgNet presented in Algorithm~\ref{alg:mgnet} and those for the FNO in Algorithm~\ref{alg:fno}. The only difference is that EV-MgNet takes the smallest number of Fourier modes on different grids, denoted as $k^\ell_{\max}$ for $\ell=1:J$. Typically, we decrease $k^\ell_{\max}$ as the grid index $\ell$ increases.
Figure~\ref{fig:Fourier-error} demonstrates that EV-MgNet can better capture the low-frequency features of the real solution than the standard V-MgNet.

\section{Numerical experiments}\label{sec:numerics}
In this section, we compare our proposed enhanced V-MgNet~(EV-MgNet) operator with several PDE data sets \cite{li2020fourier,gupta2021multiwavelet}, such as the 1D Burgers', the 1D Korteweg-de Vries (KdV), the 2D Darcy flow, and the 2D Navier-Stokes equations.  
As previously mentioned, we collect the inputs and outputs as the data set $\{a_i,u_i\}_{i=1}^N$, where
$a_i$ is part of the information of the equation, such as the initial condition, the parameter function, or the source term and $u_i$ is the numerical solution of the equation given by a classical method. Unless stated otherwise, the training set size is 1000 and the test set size is 200.

ReLU~\cite{nair2010rectified} is the most common activation function for MgNet in image classification~\cite{he2021interpretive,wang2022cnns}. Herein, we apply GeLU~\cite{hendrycks2016gaussian} activation in EV-MgNet as it fits better with the low-frequency correction structure.
Our models are trained for 600 epochs using the Adam optimizer~\cite{kingma2014adam} with an initial learning rate (LR) of $0.001$. The LR decays after every 75 epochs with a factor of $\gamma=\frac{1}{2}$. The loss function is the relative $\ell^2$ error, as shown in \eqref{eq:loss}. All experiments were performed using a single Nvidia V100 32 GB GPU, and the results were averaged over three seeds.

We compare the performance of our EV-MgNet operator model with some recently proposed neural operators. Herein, we consider a graph neural operator (GNO)~\cite{anandkumar2020neural}, a low-rank neural operator (LNO)~\cite{lu2019deeponet}, a multipole graph neural operator (MGNO)~\cite{li2020multipole}, a Fourier neural operator (FNO)~\cite{li2020fourier}, a Galerkin transformer (GT)~\cite{cao2021choose}, a multiwavelet-based transformer (MWT) \cite{gupta2021multiwavelet} with two orthogonal bases (Leg and Cheb), ResNet~\cite{he2016deep}, a classical image regression convolution model U-Net~\cite{ronneberger2015u}, and a spatial and temporal CNN for learning turbulent flows TF-Net~\cite{wang2020towards} as the benchmarks. 

\subsection{1D Burgers' equation}
We first consider the following Burgers' equation in one space dimension, which plays an important role in fluid mechanics, nonlinear acoustics, gas dynamics, and traffic flows.
\begin{equation}
	\begin{cases}
		&\partial_{t} u(x, t)+\partial_{x}\left(u^{2}(x, t) / 2\right) =\nu \partial_{x x} u(x, t),~~~~~~x \in(0,1), t \in(0,1], \\
		&u(x, 0) =\mu(x), \quad x \in(0,1),\\
		&u(0,t) = u(1,t), \quad t>0.
	\end{cases}
\end{equation}

In this problem, the task is to learn the mapping from the initial condition $a = \mu(x)$ to the solutions at $u = u(x, 1)$ with periodic boundary conditions. To compare with other benchmarks, we use the Burgers' data and the results obtained in \cite{li2020fourier,gupta2021multiwavelet,cao2021choose}, and keep the same split strategy for the training and test sets. The initial condition $a=\mu(x)$ is generated using $a \sim \mathcal D$, where $\mathcal D=\mathcal{N}\left(0,625(-\Delta+25 I)^{-2}\right)$ with periodic boundary conditions. In this study, we set $\nu=0.1$ and solve this equations on a uniform grid with $2^{13}=8192$ intervals. Futher, we obtained low resolution data sets via sampling, using a uniform grid with coarser sizes.

\begin{figure}[htbp]
	\centering
	\begin{minipage}[t]{0.45\textwidth}
		\centering
		\includegraphics[width=\textwidth]{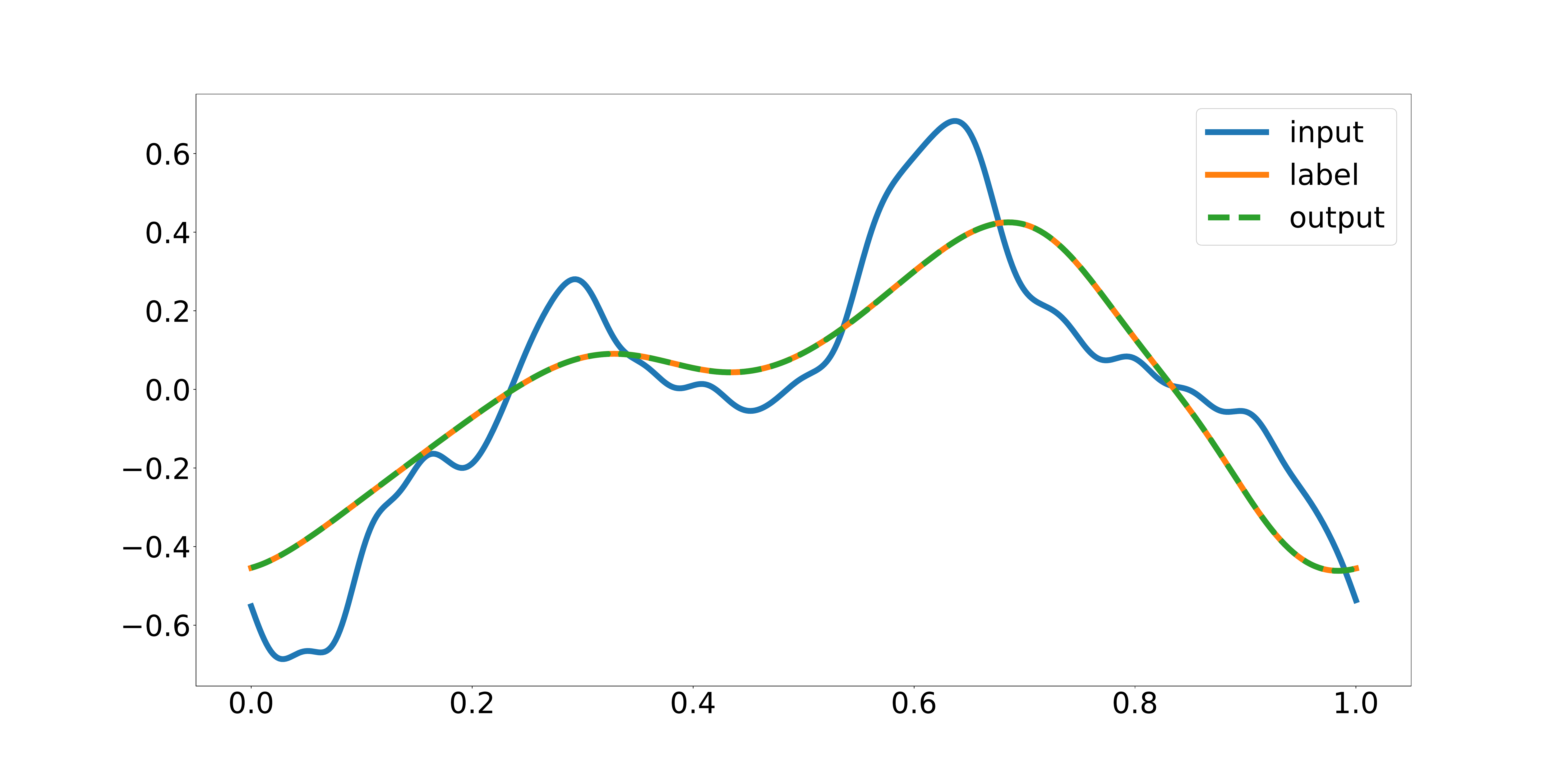}
	\end{minipage}
	\begin{minipage}[t]{0.45\textwidth}
		\centering
		\includegraphics[width=\textwidth]{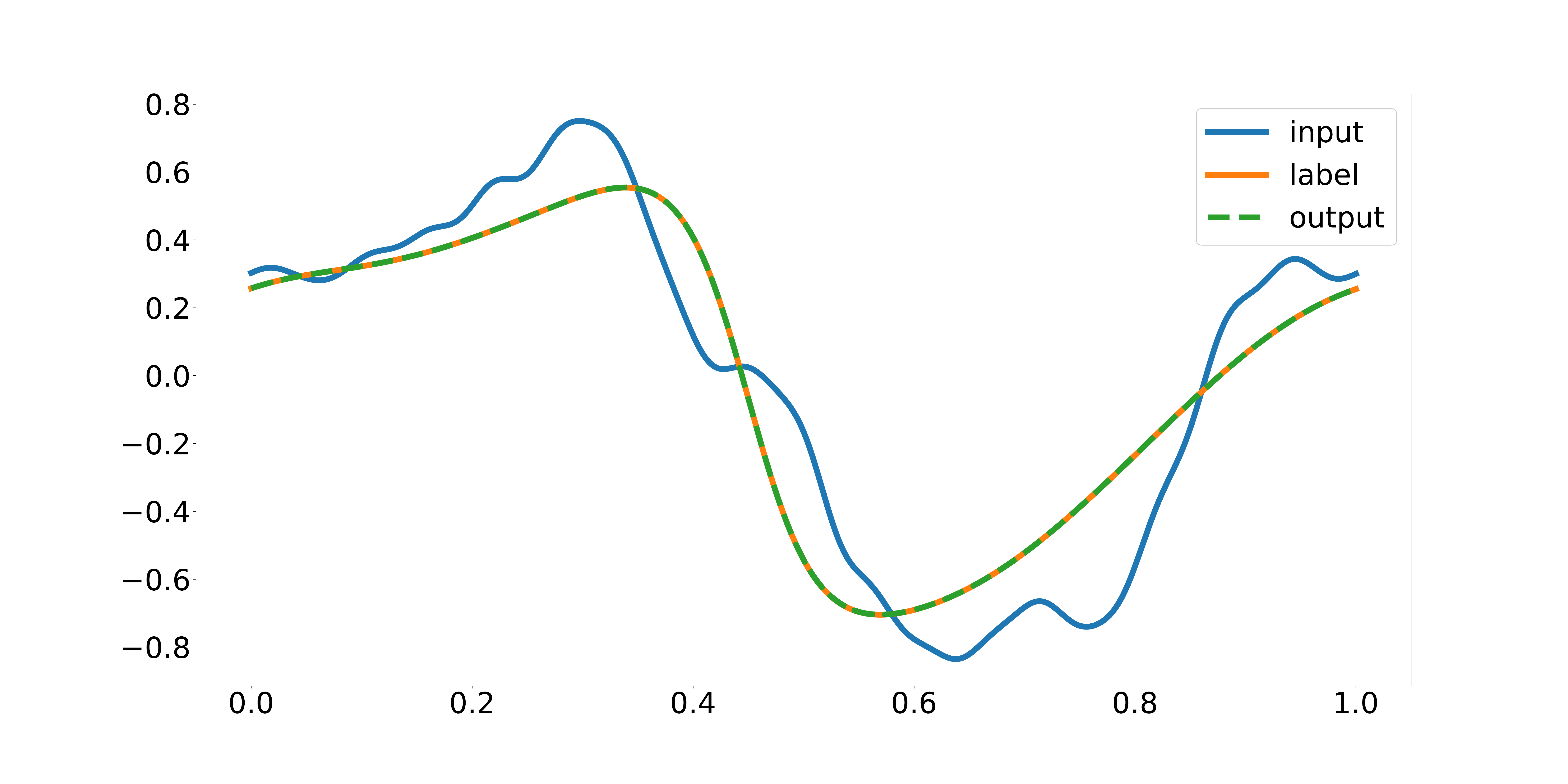}
	\end{minipage}
	\caption{Two random examples of the results predicted by our method for the 1D Burgers' equation.}
	\label{fig:burgers}
\end{figure}

The experimental results obtained using the Burger's equation for different resolutions are shown in Table \ref{results:burgers}. We show that our results have lower relative errors than other benchmarks. Figure \ref{fig:burgers} shows the predicted solution obtained using our method, and it fits the label solution well.

\begin{sidewaystable}
	\sidewaystablefn%
	\begin{center}
		\begin{minipage}{\textheight}
			\caption{Results of 1D Burger's equation evaluated at various resolutions.}\label{results:burgers}
			\begin{tabular*}{\textheight}{@{\extracolsep{\fill}}lcccccc@{\extracolsep{\fill}}}
				\toprule
				model 	      &   256   &  512   &1024    & 2048& 4096 &8192  \\
				\midrule
				FNO      &  $1.49\times10^{-2}$& $1.58\times10^{-2}$ & $1.60\times10^{-2}$ & $1.46\times10^{-2}$ &$1.42\times10^{-2}$ &$1.39\times10^{-2}$ \\ 
				
				GT      &   $1.26\times 10^{-3}$      & $1.30\times 10^{-3}$ &   $1.14\times 10^{-3}$     & $1.00\times10^{-3}$ &$1.10\times 10^{-3}$ &$1.12\times 10^{-3}$\\    
				
				MGNO         &  $2.43\times10^{-2}$& $3.55\times10^{-2}$ & $3.74\times10^{-2}$ & $3.60\times10^{-2}$ &$3.64\times10^{-2}$ &$3.64\times10^{-2}$ \\ 
				
				LNO         &  $2.12\times10^{-2}$& $2.21\times10^{-2}$ & $2.17\times10^{-2}$ & $2.19\times10^{-2}$ &$2.00\times10^{-2}$ &$1.89\times10^{-2}$ \\    
				
				GNO         &  $5.55\times10^{-2}$& $5.94\times10^{-2}$ & $6.51\times10^{-2}$ & $6.63\times10^{-2}$ &$6.66\times10^{-2}$ &$6.99\times10^{-2}$ \\    
				
				MWT Leg  &  $1.99\times10^{-3}$& $1.85\times10^{-3}$ & $1.84\times10^{-3}$ & $1.86\times10^{-3}$ &$1.85\times10^{-3}$ &$1.78\times10^{-3}$ \\  
				
				MWT Cheb  &  $4.02\times10^{-3}$& $3.81\times10^{-3}$ & $3.36\times10^{-3}$ & $3.95\times10^{-3}$ &$2.99\times10^{-3}$ &$2.89\times10^{-3}$ \\   
				
				\midrule
				V-MgNet     &   $6.06\times 10^{-2}$      & $6.43\times 10^{-2}$ &   $9.44\times 10^{-2}$     & $1.59\times10^{-1}$ &$2.54\times 10^{-1}$ &$2.97\times 10^{-1}$\\
				EV-MgNet & \textbf{$6.38\times10^{-4}$} &$6.21\times10^{-4}$ &$5.49\times10^{-4}$  &$5.74\times10^{-4}$ &$5.91\times10^{-4}$  &$5.88\times10^{-4}$\\
				\botrule
			\end{tabular*}
			
		\end{minipage}
	\end{center}
\end{sidewaystable}

\subsection{1D KDV equation}
Here, we consider the 1D Korteweg–De Vries (KdV) equation which is a mathematical model of waves on shallow water surfaces. The 1D KdV equation is a nonlinear, dispersive partial differential equation for a function of two dimensionless real variables, $x$ and $t$ which are proportional to space and time, respectively.
\begin{equation}
	\begin{cases}
		&\partial_t u(x,t) =-0.5 u(x,t)\partial_x u(x,t) -\partial_{xxx} u(x,t), \quad x \in(0,1), t \in(0,1], \\
		&u(x,0) = \mu(x), \quad x\in(0,1),\\
		&u(0,t)=u(1,t), \quad t>0.
	\end{cases}
\end{equation}

In this problem, the task is to learn the mapping from the initial condition $a = \mu(x)$ to the solutions $u=u(x, 1)$. The method generating the data follows~\cite{gupta2021multiwavelet} that  the initial condition in Gaussian random fields according to $a(x) \sim \mathcal{N}\left(0,7^{4}\left(-\Delta+7^{2} I\right)^{-2.5}\right)$ with periodic boundary conditions. 
Here, we solve the above problem using chebfun package~\cite{driscoll2014chebfun} on a uniform grid with $2^{10}=1024$ intervals, and the low-resolution data sets are obtained by sampling on a uniform grid with coarser sizes.

\begin{figure}[htbp]
	\centering
	\begin{minipage}[t]{0.48\textwidth}
		\centering
		\includegraphics[width=\textwidth]{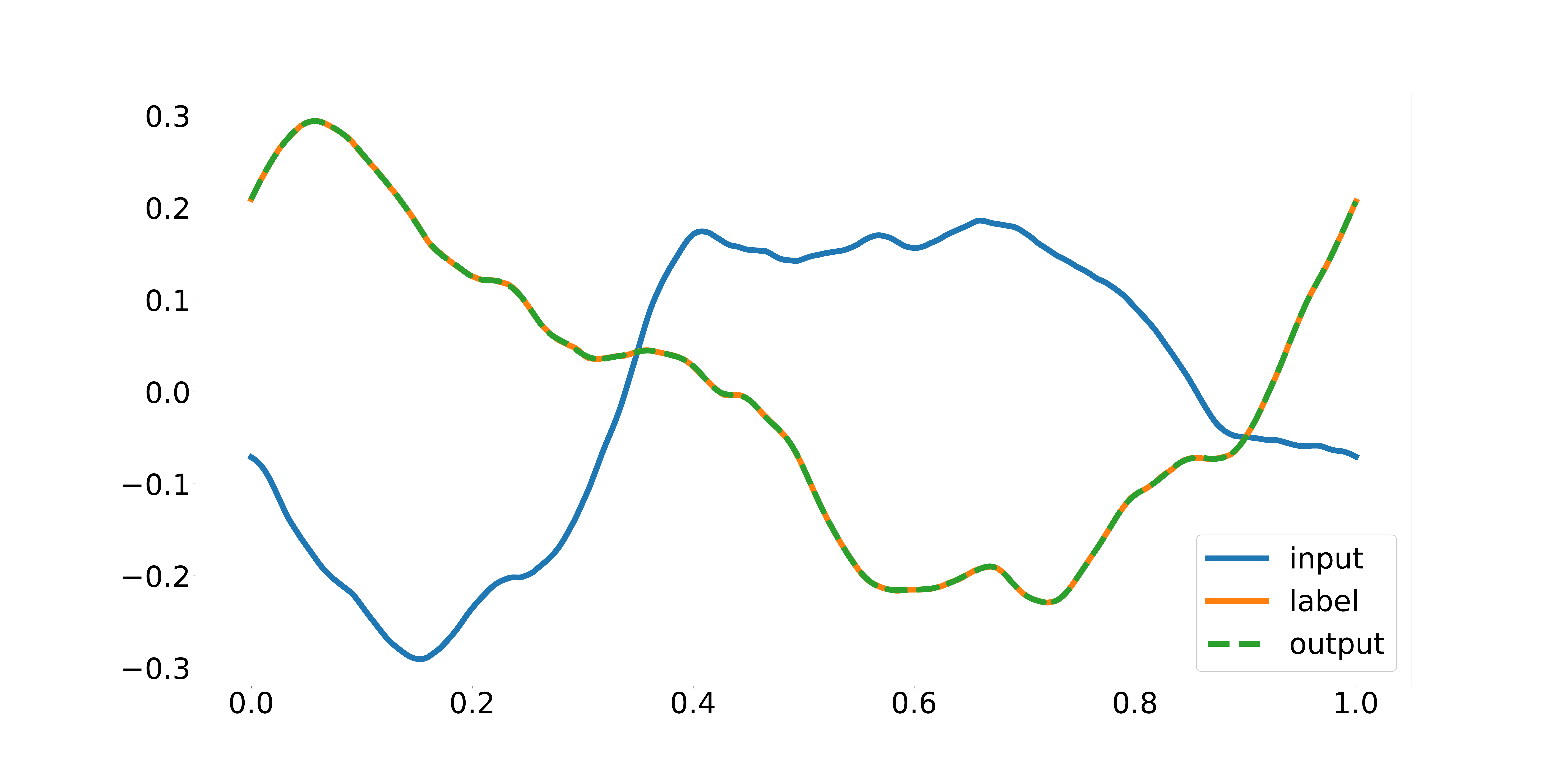}
	\end{minipage}
	\begin{minipage}[t]{0.48\textwidth}
		\centering
		\includegraphics[width=\textwidth]{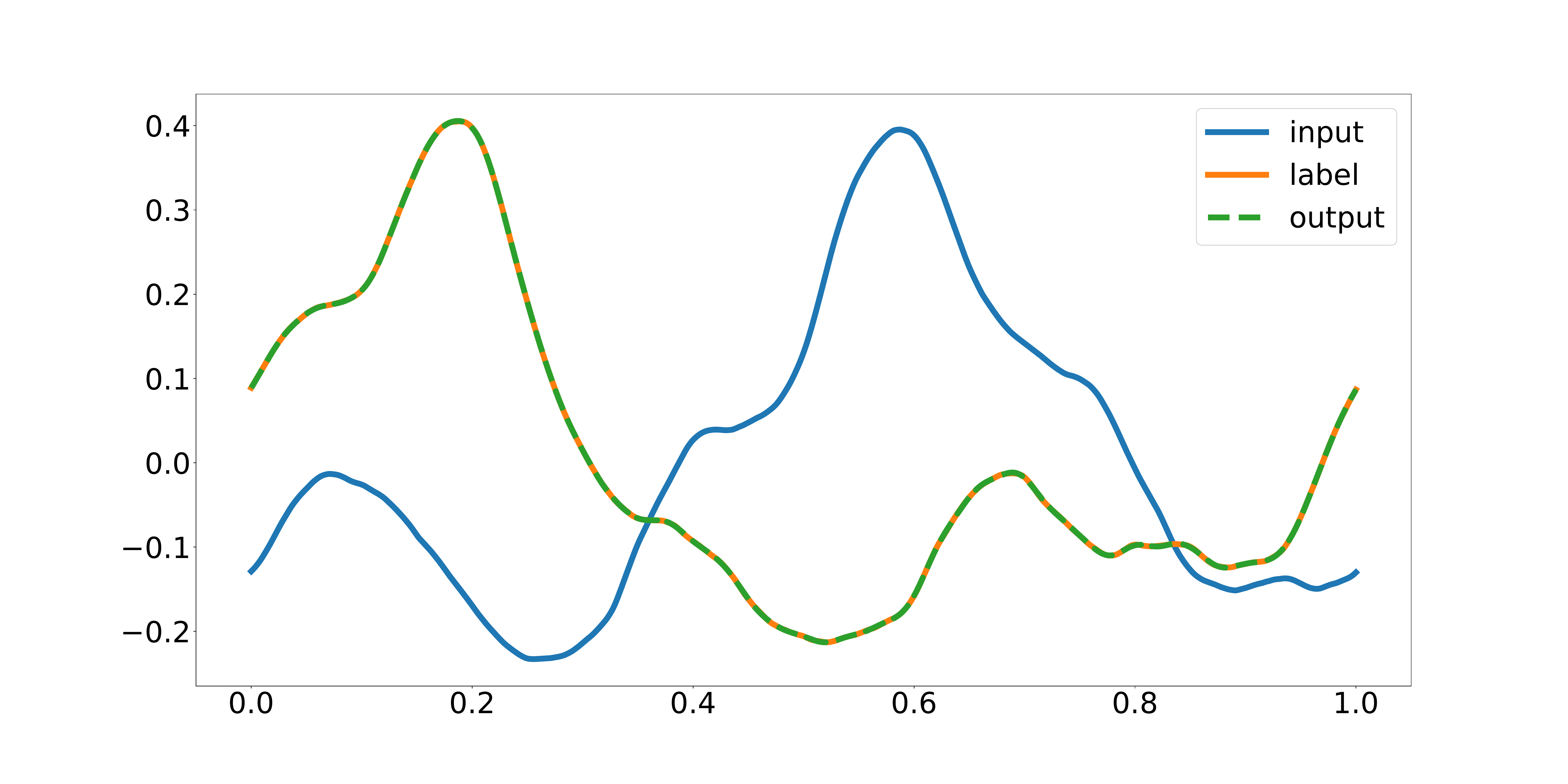}
	\end{minipage}
	\caption{Two random examples of the results predicted by our method for the 1D KdV equation.}
	\label{fig:kdv}
\end{figure}

The results of the experiments on the KdV equation for different resolutions are shown in Table \ref{results:kdv}.
Compared to other benchmarks, our method obtains the best relative error. 
Note that our method can outperform FNO by an order of magnitude, and our method can reduce the relative error by $50\%$ compared to the multiwavelet method; Figure \ref{fig:kdv} shows that our result fits well with the label.

\begin{table}[htp!]
	\begin{center}
		\caption{Results of the 1D KdV equation evaluated at various resolutions.}\label{results:kdv}
		\begin{tabular}{@{}llllll@{}}
			\toprule
			model 	      &   64   &  128   &256   & 512& 1024  \\
			\midrule
			FNO      &  $1.25\times 10^{-2}$ & $1.24\times 10^{-2}$ & $1.25\times 10^{-2}$ & $1.22\times 10^{-2}$&$1.26\times 10^{-2}$\\  
			
			MGNO         &  $1.30\times10^{-1}$& $1.52\times10^{-1}$ & $1.36\times10^{-1}$ & $1.35\times10^{-1}$ &$1.36\times10^{-1}$  \\ 
			
			LNO         &  $4.29\times10^{-2}$& $5.57\times10^{-2}$ & $4.14\times10^{-2}$ & $4.25\times10^{-2}$ &$4.47\times10^{-2}$ \\    
			
			GNO         &  $7.89\times10^{-2}$& $7.60\times10^{-2}$ & $6.95\times10^{-2}$ & $6.99\times10^{-2}$ &$7.21\times10^{-2}$\\    
			
			MWT Leg  &  $3.38\times10^{-3}$& $3.75\times10^{-3}$ & $4.18\times10^{-3}$ & $3.93\times10^{-3}$ &$3.89\times10^{-3}$ \\  
			
			MWT Cheb  &  $7.15\times10^{-3}$& $7.12\times10^{-3}$ & $6.04\times10^{-3}$ & $7.69\times10^{-3}$ &$6.75\times10^{-3}$ \\   
			
			\midrule
			V-MgNet  & \textbf{$3.45\times10^{-1}$} &$4.99\times10^{-1}$ &$5.90\times10^{-1}$  &$6.34\times10^{-1}$ &$6.63\times10^{-1}$ \\
			EV-MgNet & \textbf{$2.49\times10^{-3}$} &$8.66\times10^{-4}$ &$1.75\times10^{-3}$  &$1.75\times10^{-3}$ &$1.75\times10^{-3}$ \\
			\botrule
		\end{tabular}
	\end{center}
	
\end{table}

\subsection{2D Darcy flow}
\label{darcy}
Darcy's law is a second-order linear elliptic equation that	describes the flow of a fluid through a porous medium with the following form:
\begin{equation}
	\begin{cases}
		&-\nabla \cdot(a(x) \nabla u(x)) =f(x), \quad x \in(0,1)^{2}, \\
		&u(x) =0, \quad x \in \partial(0,1)^{2}.
	\end{cases}
\end{equation}
In this problem, we aim to learn the operator from the diffusion coefficient $a(x)$ to the solution $u(x)$ with the forcing function $f(x)=1$. The diffusion coefficients are generated to $a(x) \sim \mathcal D$, where $\mathcal D=\mathcal{N}\left(0,(-\Delta+9 I)^{-2}\right)$, and $\Delta$ is the Laplacian with zero Neumann boundary conditions, and mapping $\psi: \mathbb{R} \rightarrow \mathbb{R}$ takes the value 12 on the positive part of the real line and 3 on the negative, and the push-forward is defined pointwise. The original solutions $u(x)$ are obtained using a second-order finite difference scheme on a $421 \times 421$ grid. Then, low-resolution data sets are sub-sampled from the original solutions.

\begin{table}[htp!]
	\begin{center}
		\caption{Results of the 2D Darcy flow equation}\label{results:darcy}
		\begin{tabular}{@{}lllll@{}}
			\toprule
			Model 	      &   $85\times85$   &  $141\times141$   &$211\times211$    & $421\times421$  \\
			\midrule
			GNO              &  $3.46\times 10^{-2}$ & $3.32\times 10^{-2}$ & $3.42\times 10^{-2}$ & $3.69\times 10^{-2}$ \\
			LNO              &  $5.20\times 10^{-2}$ & $4.61\times 10^{-2}$ & $4.45\times 10^{-2}$ & -\\
			MGNO           &  $4.16\times 10^{-2}$ & $4.28\times 10^{-2}$ & $4.28\times 10^{-2}$ & $4.20\times 10^{-2}$ \\
			FNO           &  $1.08\times 10^{-2}$ & $1.09\times 10^{-2}$ & $1.09\times 10^{-2}$ & $0.98\times 10^{-2}$ \\
			GT      &  $8.51\times 10^{-3}$       & $8.40\times 10^{-3}$ & $8.50\times 10^{-3}$       & $8.93\times 10^{-3}$         \\
			MWT Leg  &  $8.54\times 10^{-3}$ &  $7.32\times 10^{-3}$&  $7.27\times 10^{-3}$ &  $6.86\times 10^{-3}$ \\
			MWT Cheb  &  $9.43\times 10^{-3}$ &  $8.28\times 10^{-3}$&  $8.83\times 10^{-3}$ &  $8.74\times 10^{-3}$ \\
			\midrule
			V-MgNet              &  $5.07\times 10^{-3}$  &$1.07\times 10^{-2}$ &$2.63\times 10^{-2}$ & $5.16\times 10^{-2}$ \\
			EV-MgNet         &  $4.57\times 10^{-3}$  &$3.79\times 10^{-3}$ &$3.79\times 10^{-3}$ & $3.83\times 10^{-3}$ \\
			\botrule
		\end{tabular}
	\end{center}
\end{table}

\begin{figure}
	\centerline{\includegraphics[width=.6\textheight]{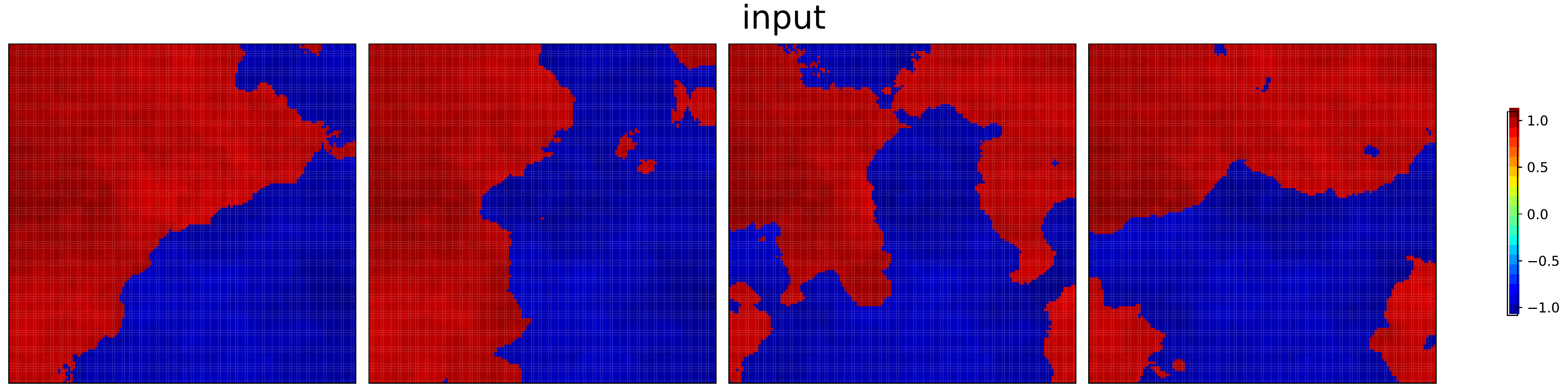}}
	\centerline{\includegraphics[width=.6\textheight]{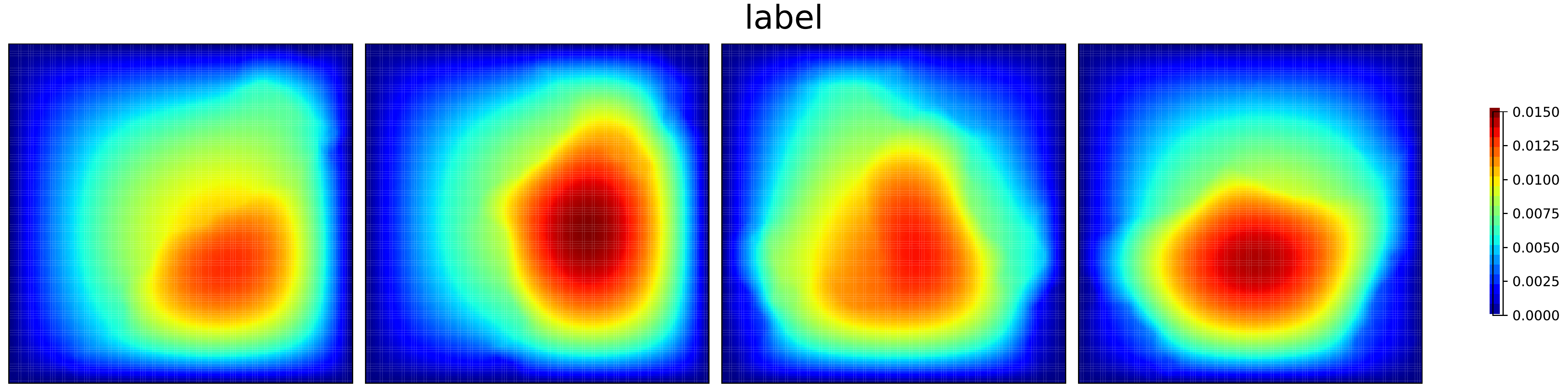}}
	\centerline{\includegraphics[width=.6\textheight]{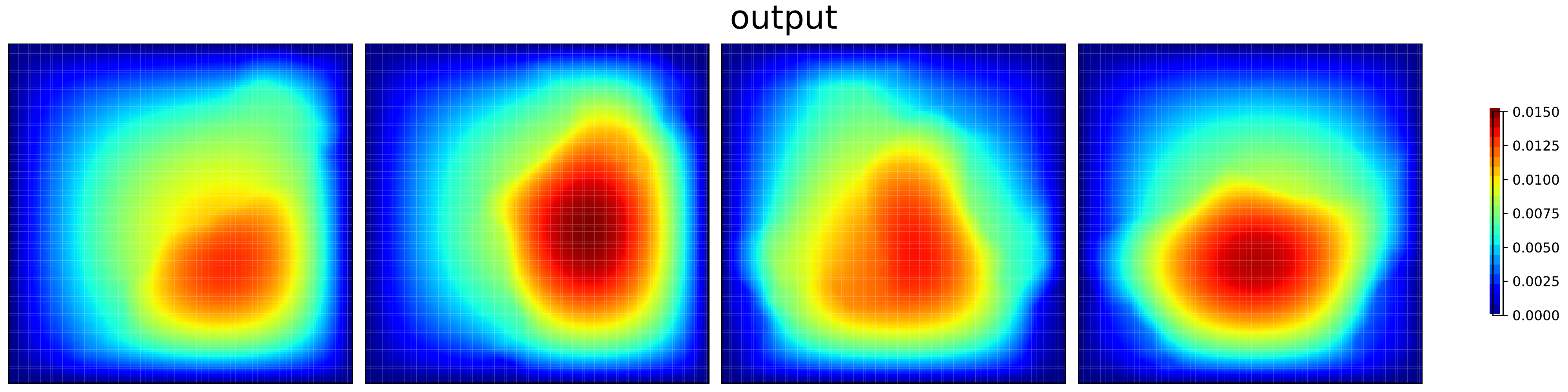}}
	\caption{A random example of the results predicted by our method for the 2D Darcy flow equation.}
	\label{fig:darcy}
\end{figure}

Table \ref{results:darcy} shows the results of the experiments on the 2D Darcy flow equation at different resolutions.  Compared with other benchmarks, our method obtains the lowest relative error. 
Our predicted solutions show that our result fits well with the label, as shown in Figure \ref{fig:darcy}.

\subsection{2D Navier-Stokes equations}
The Navier–Stokes equations are a set of PDEs that describe the motion of viscous fluids. Here, we consider the 2D Navier-Stokes equations for a viscous and incompressible fluid in vorticity form on the unit cube with the periodic boundary condition using the following equations:
\begin{equation}
	\begin{cases}
		&\partial_{t} w(x, t)+u(x, t) \cdot \nabla w(x, t) =\nu \Delta w(x, t)+f(x), \quad x \in(0,1)^{2}, t \in(0, T], \\
		&\nabla \cdot u(x, t) =0, \quad x \in(0,1)^{2}, t \in[0, T], \\
		&w(x, 0) = \mu(x), \quad x \in(0,1)^{2}.
	\end{cases}
\end{equation}
In this problem, we aim to learn an operator that maps the first 10-time steps of vorticity to the subsequent time steps until $T$. In particular, we merge the temporal and spatial dimensions to make the input a 3D tensor with one channel, i.e., $a\in \mathbb R^{d\times d\times 10\times 1}$, and the output a 3D tensor with one channel, i.e., $u\in \mathbb R^{d\times d\times T-10\times 1}$.
The input is the initial condition $a = \mu(x)$ and generated using $a \sim \mathcal D$, where $\mathcal D=\mathcal{N}\left(0,7^{3 / 2}(-\Delta+49 I)^{-2.5}\right)$ with periodic boundary conditions. The forcing is kept fixed as $f(x)=0.1\left(\sin \left(2 \pi\left(x_{1}+x_{2}\right)\right)+\right. \left.\cos \left(2 \pi\left(x_{1}+x_{2}\right)\right)\right)$. The equation is solved using the stream-function formulation and a pseudospectral method. The experiments are conducted with (1) the viscosities $\nu=10^{-3}$, final time $T=50$, number of training pairs $N=1000$; (2) $\nu=10^{-4}$, $T=30$, $N=1000$; and (3) $\nu=10^{-5}$, $T=20$, $N=1000$. We fix the resolution as $64 \times 64$ for training and testing.

\begin{table}[htp!]
	\begin{center}
		\caption{Results of the 2D Navier-Stokes equations evaluated at various viscosities $\nu$.}\label{results:nsresults}
		\begin{tabular}{@{}llll@{}}
			\toprule
			\multirow{3}[2]{*}{Networks}  	      &   $\nu=1e-3$   &$\nu=1e-4$ &$\nu=1e-5$\\
			&    T=50             & T=30                    & T=20          \\
			&  N=1000           & N=1000           & N=1000    \\ 
			\midrule
			U-Net                  &  $2.45 \times 10^{-2}$ &$2.05\times 10^{-1}$   & $1.98\times 10^{-1}$ \\
			TF-Net                &  $2.25 \times 10^{-2}$ &$2.25\times 10^{-1}$   & $2.27\times 10^{-1}$ \\
			ResNet                &  $7.01 \times 10^{-2}$ &$2.87\times 10^{-1}$   & $2.75\times 10^{-1}$ \\
			
			FNO-3D              &  $8.60 \times 10^{-3}$ &$1.92\times 10^{-1}$   & $1.89\times 10^{-1}$ \\
			FNO-2D              &  $1.28 \times 10^{-2}$ &$1.56\times 10^{-1}$   & $1.56\times 10^{-1}$ \\
			MWT Leg            &  $6.25 \times 10^{-3}$ &$1.52\times 10^{-1}$  &   $1.54\times 10^{-1}$ \\
			MWT Cheb            &  $7.20 \times 10^{-3}$ &$1.57\times 10^{-1}$  &   $1.67\times 10^{-1}$ \\
			\midrule
			V-MgNet           &  $2.53 \times 10^{-1}$ &$6.84\times 10^{-1}$   &   $7.57\times 10^{-1}$ \\
			EV-MgNet          &  $6.03 \times 10^{-3}$ &$1.42\times 10^{-1}$   &   $1.45\times 10^{-1}$ \\
			\botrule
		\end{tabular}
	\end{center}
\end{table}

\begin{figure}
	\centerline{\includegraphics[width=.6\textheight]{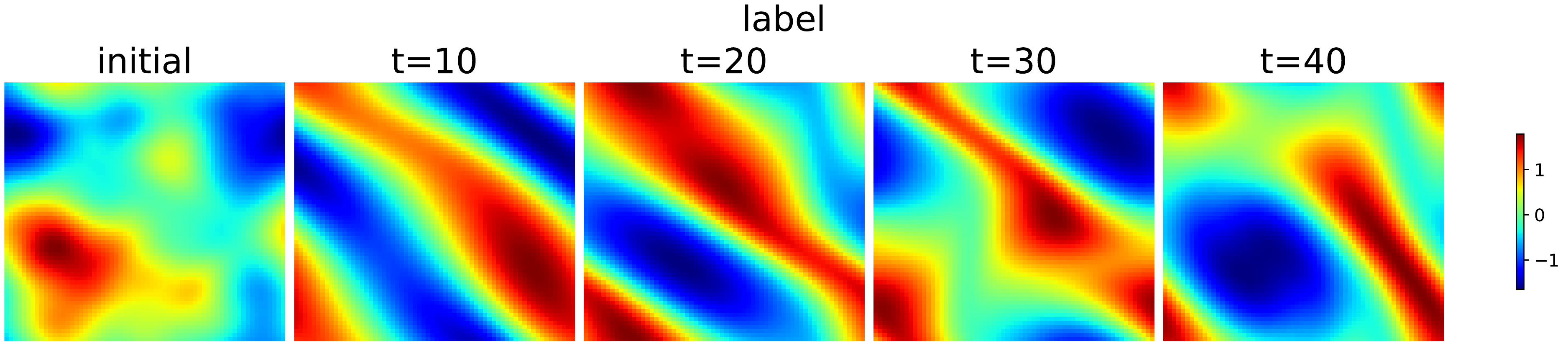}}
	\centerline{\includegraphics[width=.6\textheight]{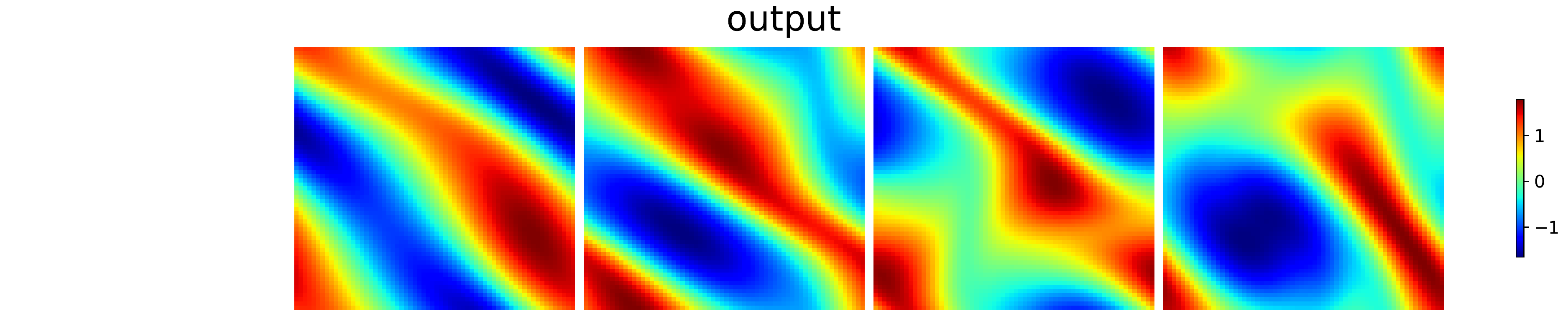}}
	\caption{A random example of the results predicted by our method for the 2D Navier-Stokes equations. In this figure, the inputs are the vorticities of the first 10-time steps, and the predicted outputs are the vorticities of the subsequent 40-time steps, i.e. $T=50$.}
	\label{fig:ns}
\end{figure}

The experimental results obtained using the 2D Navier–Stokes equations for different $\nu$ values are shown in Table \ref{results:nsresults}.
Compared to other benchmarks, our method obtains the lowest relative error.
As shown in Figure \ref{fig:ns}, our method can well predict the solution.

\subsection{Training and prediction at different resolutions}
\begin{figure}[htbp]
	\centering
	\begin{minipage}[t]{0.48\textwidth}
		\centering
		\includegraphics[width=\textwidth]{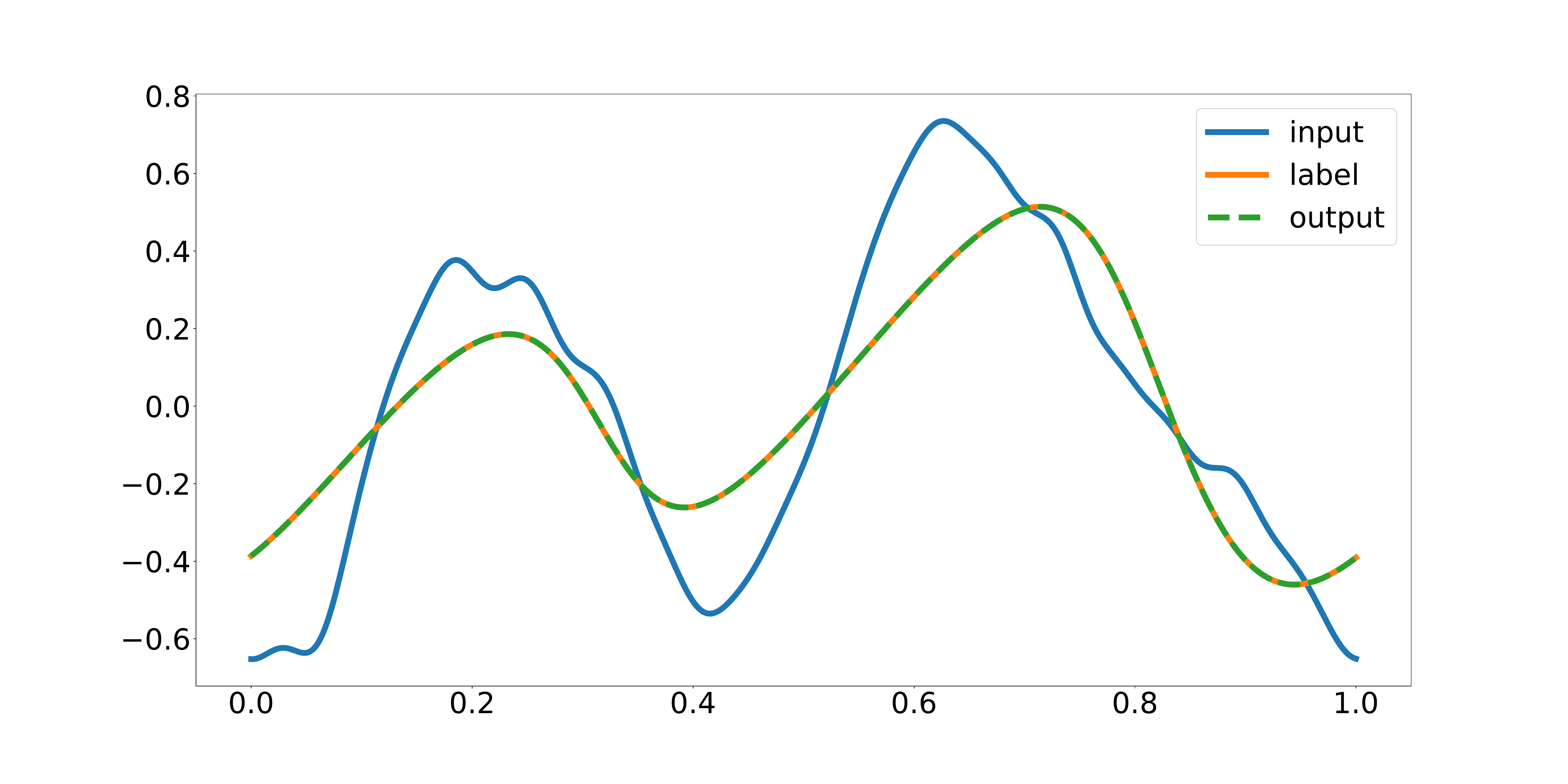}
	\end{minipage}
	\begin{minipage}[t]{0.48\textwidth}
		\centering
		\includegraphics[width=\textwidth]{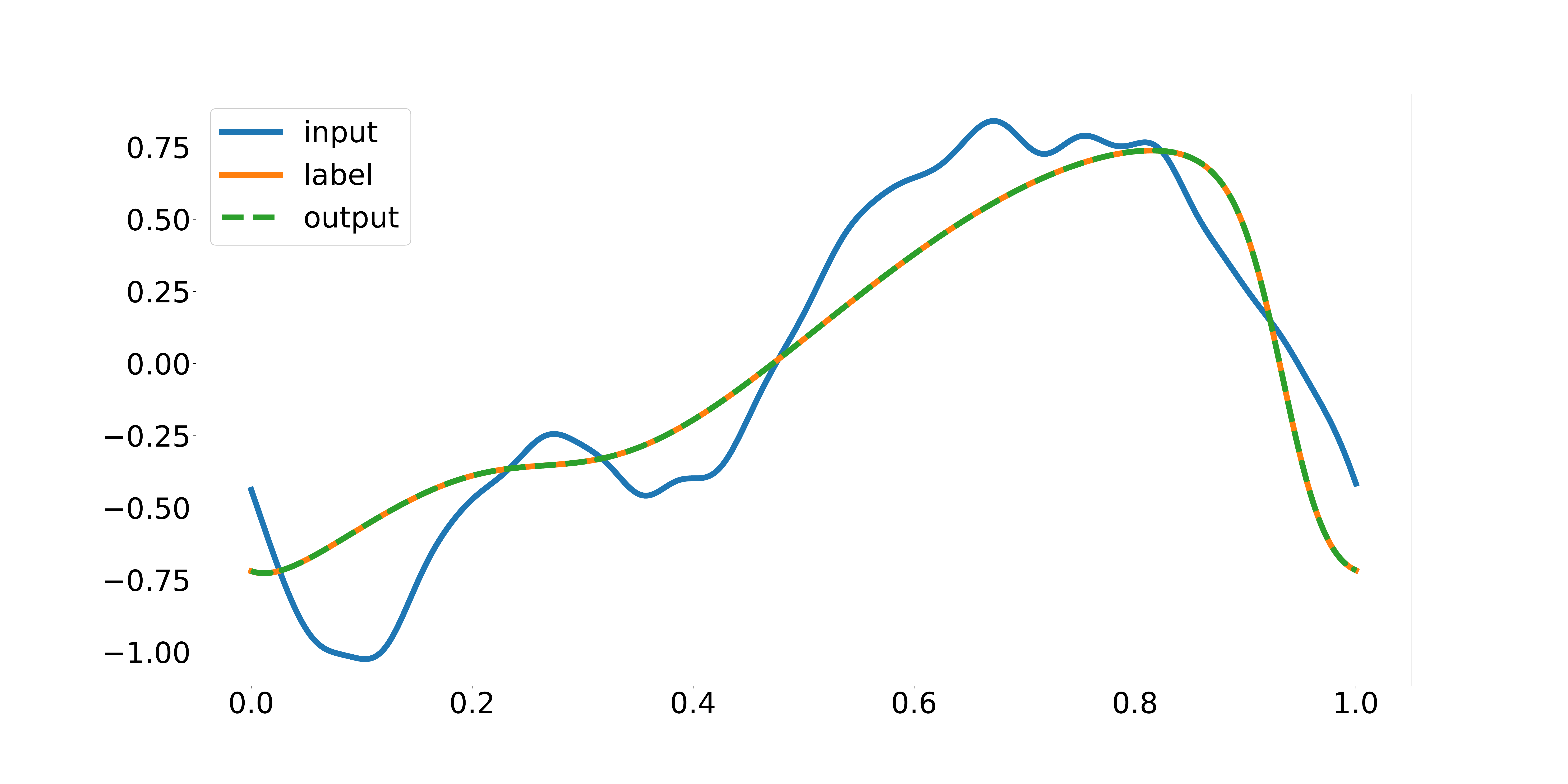}
	\end{minipage}
	\caption{Two random examples of training and testing at 512 and 8192 resolutions, respectively.}
	\label{fig:burgersresolution}
\end{figure}

One crucial feature is that the input $a$ and the output ${\rm EV-MgNet}(a)$ have the same spatial dimension because EV-MgNet is mainly based on convolution operations. 
Thus, the EV-MgNet can be trained at a lower computational cost on data sets with lower resolutions but tested with data sets with higher resolutions.  
By using the 1D Burgers' equation as an example, Tables \ref{results:multivalet-burgers-resolution} and \ref{results:mgnet-burgers-resolution} show the results of 
A multiwavelet-based transfomer with Legendre orthogonal bases (MWT Leg)~\cite{gupta2021multiwavelet}
and EV-MgNet operator learning models at low resolutions ($d=128, 256, 512$) and predicting at high resolutions ($d=2048, 4096, 8192$), respectively. 
However, the EV-MgNet has (10 times) lower relative errors than the MWT Leg model. In addition, the EV-MgNet model produces more stable results in training at a fixed resolution but testing at different resolutions. These two phenomena demonstrate that our model is more robust in training on low-resolution data than in testing on high-resolution data.
Figure \ref{fig:burgersresolution} shows two examples of training and testing at 512 and 8192 resolutions, respectively; the predicted solutions are close to the label solutions. 

\begin{table}[htp!]
	\begin{center}
		\caption{MWT Leg model trained at lower resolutions and predicted at higher resolutions}\label{results:multivalet-burgers-resolution}
		\begin{tabular}{cccc}
			\toprule
			\diagbox{Train}{Test} 	    & 2048& 4096 &8192  \\
			\midrule
			128   &$3.68\times10^{-2}$ &$3.89\times10^{-2}$  &$4.56\times10^{-2}$\\
			256    &$2.26\times10^{-2}$ &$2.81\times10^{-2}$  &$3.21\times10^{-2}$\\
			512  & $1.40\times10^{-2}$ &$1.91\times10^{-2}$  &$2.41\times10^{-2}$ \\   
			\botrule
		\end{tabular}
	\end{center}

\end{table}

\begin{table}[htp!]
	\begin{center}
		\caption{EV-MgNet operator model trained at low resolutions and predicted at high resolutions}\label{results:mgnet-burgers-resolution}
		\begin{tabular}{cccc}
			\toprule
			\diagbox{Train}{Test} 	    & 2048& 4096 &8192  \\
			\midrule
			128   &$3.99\times10^{-3}$ &$4.11\times10^{-3}$  &$4.19\times10^{-3}$\\
			256    &$8.93\times10^{-4}$ &$9.26\times10^{-4}$  &$9.39\times10^{-4}$\\
			512  & $6.46\times10^{-4}$ &$6.56\times10^{-4}$  &$6.62\times10^{-4}$ \\   
			\botrule
		\end{tabular}
	\end{center}
\end{table}

\section{Concluding remarks}\label{sec:conclusion}
In this study, we propose EV-MgNet, an MgNet-based operator learning architecture for solving numerical PDEs. 
With the enhancement of low-frequency correction, our new model can  handle the high- and low-frequency components of errors simultaneously during standard V-MgNet iterations (between layers). 
Based on extensive experiments, our model exhibits a low relative error and good prediction accuracy at different resolutions.

In the future, we plan to introduce more general and intrinsic enhancement techniques into V-MgNet from the MG perspective for faster and more accurate operator learning architecture. In addition, it is important to theoretically and experimentally study the decay rates of the frequencies in the iteration processes of V-MgNet and EV-MgNet. 
Naturally, more data sets of operator learning in the case of PDEs can and should be used with our model.


\bibliography{sn-article.bib}


\begin{thebibliography}{44}
\ifx \bisbn   \undefined \def \bisbn  #1{ISBN #1}\fi
\ifx \binits  \undefined \def \binits#1{#1}\fi
\ifx \bauthor  \undefined \def \bauthor#1{#1}\fi
\ifx \batitle  \undefined \def \batitle#1{#1}\fi
\ifx \bjtitle  \undefined \def \bjtitle#1{#1}\fi
\ifx \bvolume  \undefined \def \bvolume#1{\textbf{#1}}\fi
\ifx \byear  \undefined \def \byear#1{#1}\fi
\ifx \bissue  \undefined \def \bissue#1{#1}\fi
\ifx \bfpage  \undefined \def \bfpage#1{#1}\fi
\ifx \blpage  \undefined \def \blpage #1{#1}\fi
\ifx \burl  \undefined \def \burl#1{\textsf{#1}}\fi
\ifx \doiurl  \undefined \def \doiurl#1{\url{https://doi.org/#1}}\fi
\ifx \betal  \undefined \def \betal{\textit{et al.}}\fi
\ifx \binstitute  \undefined \def \binstitute#1{#1}\fi
\ifx \binstitutionaled  \undefined \def \binstitutionaled#1{#1}\fi
\ifx \bctitle  \undefined \def \bctitle#1{#1}\fi
\ifx \beditor  \undefined \def \beditor#1{#1}\fi
\ifx \bpublisher  \undefined \def \bpublisher#1{#1}\fi
\ifx \bbtitle  \undefined \def \bbtitle#1{#1}\fi
\ifx \bedition  \undefined \def \bedition#1{#1}\fi
\ifx \bseriesno  \undefined \def \bseriesno#1{#1}\fi
\ifx \blocation  \undefined \def \blocation#1{#1}\fi
\ifx \bsertitle  \undefined \def \bsertitle#1{#1}\fi
\ifx \bsnm \undefined \def \bsnm#1{#1}\fi
\ifx \bsuffix \undefined \def \bsuffix#1{#1}\fi
\ifx \bparticle \undefined \def \bparticle#1{#1}\fi
\ifx \barticle \undefined \def \barticle#1{#1}\fi
\bibcommenthead
\ifx \bconfdate \undefined \def \bconfdate #1{#1}\fi
\ifx \botherref \undefined \def \botherref #1{#1}\fi
\ifx \url \undefined \def \url#1{\textsf{#1}}\fi
\ifx \bchapter \undefined \def \bchapter#1{#1}\fi
\ifx \bbook \undefined \def \bbook#1{#1}\fi
\ifx \bcomment \undefined \def \bcomment#1{#1}\fi
\ifx \oauthor \undefined \def \oauthor#1{#1}\fi
\ifx \citeauthoryear \undefined \def \citeauthoryear#1{#1}\fi
\ifx \endbibitem  \undefined \def \endbibitem {}\fi
\ifx \bconflocation  \undefined \def \bconflocation#1{#1}\fi
\ifx \arxivurl  \undefined \def \arxivurl#1{\textsf{#1}}\fi
\csname PreBibitemsHook\endcsname

\bibitem{courant2008methods}
\begin{bbook}
\bauthor{\bsnm{Courant}, \binits{R.}},
\bauthor{\bsnm{Hilbert}, \binits{D.}}:
\bbtitle{Methods of Mathematical Physics: Partial Differential Equations}.
\bpublisher{John Wiley \& Sons},
\blocation{New York}
(\byear{2008})
\end{bbook}
\endbibitem

\bibitem{brenner2008mathematical}
\begin{bbook}
\bauthor{\bsnm{Brenner}, \binits{S.C.}},
\bauthor{\bsnm{Scott}, \binits{L.R.}},
\bauthor{\bsnm{Scott}, \binits{L.R.}}:
\bbtitle{The Mathematical Theory of Finite Element Methods}
vol. \bseriesno{3}.
\bpublisher{Springer},
\blocation{New York}
(\byear{2008})
\end{bbook}
\endbibitem

\bibitem{leveque2007finite}
\begin{bbook}
\bauthor{\bsnm{LeVeque}, \binits{R.J.}}:
\bbtitle{Finite Difference Methods for Ordinary and Partial Differential
  Equations: Steady-state and Time-dependent Problems}.
\bpublisher{SIAM},
\blocation{Philadelphia}
(\byear{2007})
\end{bbook}
\endbibitem

\bibitem{lu2019deeponet}
\begin{botherref}
\oauthor{\bsnm{Lu}, \binits{L.}},
\oauthor{\bsnm{Jin}, \binits{P.}},
\oauthor{\bsnm{Karniadakis}, \binits{G.E.}}:
Deeponet: Learning nonlinear operators for identifying differential equations
  based on the universal approximation theorem of operators.
arXiv preprint arXiv:1910.03193
(2019)
\end{botherref}
\endbibitem

\bibitem{li2020fourier}
\begin{botherref}
\oauthor{\bsnm{Li}, \binits{Z.}},
\oauthor{\bsnm{Kovachki}, \binits{N.}},
\oauthor{\bsnm{Azizzadenesheli}, \binits{K.}},
\oauthor{\bsnm{Liu}, \binits{B.}},
\oauthor{\bsnm{Bhattacharya}, \binits{K.}},
\oauthor{\bsnm{Stuart}, \binits{A.}},
\oauthor{\bsnm{Anandkumar}, \binits{A.}}:
Fourier neural operator for parametric partial differential equations.
arXiv preprint arXiv:2010.08895
(2020)
\end{botherref}
\endbibitem

\bibitem{nelsen2021random}
\begin{barticle}
\bauthor{\bsnm{Nelsen}, \binits{N.H.}},
\bauthor{\bsnm{Stuart}, \binits{A.M.}}:
\batitle{The random feature model for input-output maps between banach spaces}.
\bjtitle{SIAM Journal on Scientific Computing}
\bvolume{43}(\bissue{5}),
\bfpage{3212}--\blpage{3243}
(\byear{2021})
\end{barticle}
\endbibitem

\bibitem{gupta2021multiwavelet}
\begin{botherref}
\oauthor{\bsnm{Gupta}, \binits{G.}},
\oauthor{\bsnm{Xiao}, \binits{X.}},
\oauthor{\bsnm{Bogdan}, \binits{P.}}:
Multiwavelet-based operator learning for differential equations.
Advances in Neural Information Processing Systems
\textbf{34}
(2021)
\end{botherref}
\endbibitem

\bibitem{lecun2015deep}
\begin{barticle}
\bauthor{\bsnm{LeCun}, \binits{Y.}},
\bauthor{\bsnm{Bengio}, \binits{Y.}},
\bauthor{\bsnm{Hinton}, \binits{G.}}:
\batitle{Deep learning}.
\bjtitle{nature}
\bvolume{521}(\bissue{7553}),
\bfpage{436}--\blpage{444}
(\byear{2015})
\end{barticle}
\endbibitem

\bibitem{he2016deep}
\begin{bchapter}
\bauthor{\bsnm{He}, \binits{K.}},
\bauthor{\bsnm{Zhang}, \binits{X.}},
\bauthor{\bsnm{Ren}, \binits{S.}},
\bauthor{\bsnm{Sun}, \binits{J.}}:
\bctitle{Deep residual learning for image recognition}.
In: \bbtitle{Proceedings of the IEEE Conference on Computer Vision and Pattern
  Recognition},
pp. \bfpage{770}--\blpage{778}
(\byear{2016})
\end{bchapter}
\endbibitem

\bibitem{vaswani2017attention}
\begin{botherref}
\oauthor{\bsnm{Vaswani}, \binits{A.}},
\oauthor{\bsnm{Shazeer}, \binits{N.}},
\oauthor{\bsnm{Parmar}, \binits{N.}},
\oauthor{\bsnm{Uszkoreit}, \binits{J.}},
\oauthor{\bsnm{Jones}, \binits{L.}},
\oauthor{\bsnm{Gomez}, \binits{A.N.}},
\oauthor{\bsnm{Kaiser}, \binits{{\L}.}},
\oauthor{\bsnm{Polosukhin}, \binits{I.}}:
Attention is all you need.
Advances in neural information processing systems
\textbf{30}
(2017)
\end{botherref}
\endbibitem

\bibitem{silver2016mastering}
\begin{barticle}
\bauthor{\bsnm{Silver}, \binits{D.}},
\bauthor{\bsnm{Huang}, \binits{A.}},
\bauthor{\bsnm{Maddison}, \binits{C.J.}},
\bauthor{\bsnm{Guez}, \binits{A.}},
\bauthor{\bsnm{Sifre}, \binits{L.}},
\bauthor{\bsnm{Van Den~Driessche}, \binits{G.}},
\bauthor{\bsnm{Schrittwieser}, \binits{J.}},
\bauthor{\bsnm{Antonoglou}, \binits{I.}},
\bauthor{\bsnm{Panneershelvam}, \binits{V.}},
\bauthor{\bsnm{Lanctot}, \binits{M.}}, \betal:
\batitle{Mastering the game of go with deep neural networks and tree search}.
\bjtitle{nature}
\bvolume{529}(\bissue{7587}),
\bfpage{484}--\blpage{489}
(\byear{2016})
\end{barticle}
\endbibitem

\bibitem{guo2016convolutional}
\begin{bchapter}
\bauthor{\bsnm{Guo}, \binits{X.}},
\bauthor{\bsnm{Li}, \binits{W.}},
\bauthor{\bsnm{Iorio}, \binits{F.}}:
\bctitle{Convolutional neural networks for steady flow approximation}.
In: \bbtitle{Proceedings of the 22nd ACM SIGKDD International Conference on
  Knowledge Discovery and Data Mining},
pp. \bfpage{481}--\blpage{490}
(\byear{2016})
\end{bchapter}
\endbibitem

\bibitem{zhu2018bayesian}
\begin{barticle}
\bauthor{\bsnm{Zhu}, \binits{Y.}},
\bauthor{\bsnm{Zabaras}, \binits{N.}}:
\batitle{Bayesian deep convolutional encoder--decoder networks for surrogate
  modeling and uncertainty quantification}.
\bjtitle{Journal of Computational Physics}
\bvolume{366},
\bfpage{415}--\blpage{447}
(\byear{2018})
\end{barticle}
\endbibitem

\bibitem{adler2017solving}
\begin{barticle}
\bauthor{\bsnm{Adler}, \binits{J.}},
\bauthor{\bsnm{{\"O}ktem}, \binits{O.}}:
\batitle{Solving ill-posed inverse problems using iterative deep neural
  networks}.
\bjtitle{Inverse Problems}
\bvolume{33}(\bissue{12}),
\bfpage{124007}
(\byear{2017})
\end{barticle}
\endbibitem

\bibitem{bhatnagar2019prediction}
\begin{barticle}
\bauthor{\bsnm{Bhatnagar}, \binits{S.}},
\bauthor{\bsnm{Afshar}, \binits{Y.}},
\bauthor{\bsnm{Pan}, \binits{S.}},
\bauthor{\bsnm{Duraisamy}, \binits{K.}},
\bauthor{\bsnm{Kaushik}, \binits{S.}}:
\batitle{Prediction of aerodynamic flow fields using convolutional neural
  networks}.
\bjtitle{Computational Mechanics}
\bvolume{64}(\bissue{2}),
\bfpage{525}--\blpage{545}
(\byear{2019})
\end{barticle}
\endbibitem

\bibitem{khoo2021solving}
\begin{barticle}
\bauthor{\bsnm{Khoo}, \binits{Y.}},
\bauthor{\bsnm{Lu}, \binits{J.}},
\bauthor{\bsnm{Ying}, \binits{L.}}:
\batitle{Solving parametric pde problems with artificial neural networks}.
\bjtitle{European Journal of Applied Mathematics}
\bvolume{32}(\bissue{3}),
\bfpage{421}--\blpage{435}
(\byear{2021})
\end{barticle}
\endbibitem

\bibitem{chen2022meta}
\begin{botherref}
\oauthor{\bsnm{Chen}, \binits{Y.}},
\oauthor{\bsnm{Dong}, \binits{B.}},
\oauthor{\bsnm{Xu}, \binits{J.}}:
Meta-mgnet: Meta multigrid networks for solving parameterized partial
  differential equations.
Journal of Computational Physics,
110996
(2022)
\end{botherref}
\endbibitem

\bibitem{kovachki2021neural}
\begin{botherref}
\oauthor{\bsnm{Kovachki}, \binits{N.}},
\oauthor{\bsnm{Li}, \binits{Z.}},
\oauthor{\bsnm{Liu}, \binits{B.}},
\oauthor{\bsnm{Azizzadenesheli}, \binits{K.}},
\oauthor{\bsnm{Bhattacharya}, \binits{K.}},
\oauthor{\bsnm{Stuart}, \binits{A.}},
\oauthor{\bsnm{Anandkumar}, \binits{A.}}:
Neural operator: Learning maps between function spaces.
arXiv preprint arXiv:2108.08481
(2021)
\end{botherref}
\endbibitem

\bibitem{bhattacharya2020model}
\begin{botherref}
\oauthor{\bsnm{Bhattacharya}, \binits{K.}},
\oauthor{\bsnm{Hosseini}, \binits{B.}},
\oauthor{\bsnm{Kovachki}, \binits{N.B.}},
\oauthor{\bsnm{Stuart}, \binits{A.M.}}:
Model reduction and neural networks for parametric pdes.
arXiv preprint arXiv:2005.03180
(2020)
\end{botherref}
\endbibitem

\bibitem{patel2021physics}
\begin{barticle}
\bauthor{\bsnm{Patel}, \binits{R.G.}},
\bauthor{\bsnm{Trask}, \binits{N.A.}},
\bauthor{\bsnm{Wood}, \binits{M.A.}},
\bauthor{\bsnm{Cyr}, \binits{E.C.}}:
\batitle{A physics-informed operator regression framework for extracting
  data-driven continuum models}.
\bjtitle{Computer Methods in Applied Mechanics and Engineering}
\bvolume{373},
\bfpage{113500}
(\byear{2021})
\end{barticle}
\endbibitem

\bibitem{anandkumar2020neural}
\begin{bchapter}
\bauthor{\bsnm{Anandkumar}, \binits{A.}},
\bauthor{\bsnm{Azizzadenesheli}, \binits{K.}},
\bauthor{\bsnm{Bhattacharya}, \binits{K.}},
\bauthor{\bsnm{Kovachki}, \binits{N.}},
\bauthor{\bsnm{Li}, \binits{Z.}},
\bauthor{\bsnm{Liu}, \binits{B.}},
\bauthor{\bsnm{Stuart}, \binits{A.}}:
\bctitle{Neural operator: Graph kernel network for partial differential
  equations}.
In: \bbtitle{ICLR 2020 Workshop on Integration of Deep Neural Models and
  Differential Equations}
(\byear{2020})
\end{bchapter}
\endbibitem

\bibitem{cao2021choose}
\begin{botherref}
\oauthor{\bsnm{Cao}, \binits{S.}}:
Choose a transformer: Fourier or galerkin.
Advances in Neural Information Processing Systems
\textbf{34}
(2021)
\end{botherref}
\endbibitem

\bibitem{liu2022ht}
\begin{botherref}
\oauthor{\bsnm{Liu}, \binits{X.}},
\oauthor{\bsnm{Xu}, \binits{B.}},
\oauthor{\bsnm{Zhang}, \binits{L.}}:
Ht-net: Hierarchical transformer based operator learning model for multiscale
  pdes.
arXiv preprint arXiv:2210.10890
(2022)
\end{botherref}
\endbibitem

\bibitem{goswami2021physics}
\begin{botherref}
\oauthor{\bsnm{Goswami}, \binits{S.}},
\oauthor{\bsnm{Yin}, \binits{M.}},
\oauthor{\bsnm{Yu}, \binits{Y.}},
\oauthor{\bsnm{Karniadakis}, \binits{G.}}:
A physics-informed variational deeponet for predicting the crack path in
  brittle materials.
arXiv preprint arXiv:2108.06905
(2021)
\end{botherref}
\endbibitem

\bibitem{di2021deeponet}
\begin{botherref}
\oauthor{\bsnm{Di~Leoni}, \binits{P.C.}},
\oauthor{\bsnm{Lu}, \binits{L.}},
\oauthor{\bsnm{Meneveau}, \binits{C.}},
\oauthor{\bsnm{Karniadakis}, \binits{G.}},
\oauthor{\bsnm{Zaki}, \binits{T.A.}}:
Deeponet prediction of linear instability waves in high-speed boundary layers.
arXiv preprint arXiv:2105.08697
(2021)
\end{botherref}
\endbibitem

\bibitem{he2019mgnet}
\begin{botherref}
\oauthor{\bsnm{He}, \binits{J.}},
\oauthor{\bsnm{Xu}, \binits{J.}}:
Mgnet: A unified framework of multigrid and convolutional neural network.
Science China Mathematics,
1--24
(2019)
\end{botherref}
\endbibitem

\bibitem{he2021interpretive}
\begin{botherref}
\oauthor{\bsnm{He}, \binits{J.}},
\oauthor{\bsnm{Xu}, \binits{J.}},
\oauthor{\bsnm{Zhang}, \binits{L.}},
\oauthor{\bsnm{Zhu}, \binits{J.}}:
An interpretive constrained linear model for resnet and mgnet.
arXiv preprint arXiv:2112.07441
(2021)
\end{botherref}
\endbibitem

\bibitem{he2016identity}
\begin{bchapter}
\bauthor{\bsnm{He}, \binits{K.}},
\bauthor{\bsnm{Zhang}, \binits{X.}},
\bauthor{\bsnm{Ren}, \binits{S.}},
\bauthor{\bsnm{Sun}, \binits{J.}}:
\bctitle{Identity mappings in deep residual networks}.
In: \bbtitle{European Conference on Computer Vision},
pp. \bfpage{630}--\blpage{645}
(\byear{2016}).
\bcomment{Springer}
\end{bchapter}
\endbibitem

\bibitem{xu1992iterative}
\begin{barticle}
\bauthor{\bsnm{Xu}, \binits{J.}}:
\batitle{Iterative methods by space decomposition and subspace correction}.
\bjtitle{SIAM review}
\bvolume{34}(\bissue{4}),
\bfpage{581}--\blpage{613}
(\byear{1992})
\end{barticle}
\endbibitem

\bibitem{xu2002method}
\begin{barticle}
\bauthor{\bsnm{Xu}, \binits{J.}},
\bauthor{\bsnm{Zikatanov}, \binits{L.}}:
\batitle{The method of alternating projections and the method of subspace
  corrections in hilbert space}.
\bjtitle{Journal of the American Mathematical Society}
\bvolume{15}(\bissue{3}),
\bfpage{573}--\blpage{597}
(\byear{2002})
\end{barticle}
\endbibitem

\bibitem{hackbusch2013multi}
\begin{bbook}
\bauthor{\bsnm{Hackbusch}, \binits{W.}}:
\bbtitle{Multi-grid Methods and Applications}
vol. \bseriesno{4}.
\bpublisher{Springer},
\blocation{New York}
(\byear{2013})
\end{bbook}
\endbibitem

\bibitem{xu2017algebraic}
\begin{barticle}
\bauthor{\bsnm{Xu}, \binits{J.}},
\bauthor{\bsnm{Zikatanov}, \binits{L.}}:
\batitle{Algebraic multigrid methods}.
\bjtitle{Acta Numerica}
\bvolume{26},
\bfpage{591}--\blpage{721}
(\byear{2017})
\end{barticle}
\endbibitem

\bibitem{kovachki2021universal}
\begin{barticle}
\bauthor{\bsnm{Kovachki}, \binits{N.}},
\bauthor{\bsnm{Lanthaler}, \binits{S.}},
\bauthor{\bsnm{Mishra}, \binits{S.}}:
\batitle{On universal approximation and error bounds for fourier neural
  operators}.
\bjtitle{Journal of Machine Learning Research}
\bvolume{22},
(\byear{2021})
\end{barticle}
\endbibitem

\bibitem{lanthaler2022error}
\begin{barticle}
\bauthor{\bsnm{Lanthaler}, \binits{S.}},
\bauthor{\bsnm{Mishra}, \binits{S.}},
\bauthor{\bsnm{Karniadakis}, \binits{G.E.}}:
\batitle{Error estimates for deeponets: A deep learning framework in infinite
  dimensions}.
\bjtitle{Transactions of Mathematics and Its Applications}
\bvolume{6}(\bissue{1}),
\bfpage{001}
(\byear{2022})
\end{barticle}
\endbibitem

\bibitem{he2022approximation}
\begin{barticle}
\bauthor{\bsnm{He}, \binits{J.}},
\bauthor{\bsnm{Li}, \binits{L.}},
\bauthor{\bsnm{Xu}, \binits{J.}}:
\batitle{Approximation properties of deep relu cnns}.
\bjtitle{Research in the Mathematical Sciences}
\bvolume{9}(\bissue{3}),
\bfpage{1}--\blpage{24}
(\byear{2022})
\end{barticle}
\endbibitem

\bibitem{wang2022cnns}
\begin{bchapter}
\bauthor{\bsnm{Wang}, \binits{J.}},
\bauthor{\bsnm{Xu}, \binits{J.}},
\bauthor{\bsnm{Zhu}, \binits{J.}}:
\bctitle{Cnns with compact activation function}.
In: \bbtitle{International Conference on Computational Science},
pp. \bfpage{319}--\blpage{327}
(\byear{2022}).
\bcomment{Springer}
\end{bchapter}
\endbibitem

\bibitem{trottenberg2000multigrid}
\begin{bbook}
\bauthor{\bsnm{Trottenberg}, \binits{U.}},
\bauthor{\bsnm{Oosterlee}, \binits{C.W.}},
\bauthor{\bsnm{Schuller}, \binits{A.}}:
\bbtitle{Multigrid}.
\bpublisher{Elsevier},
\blocation{San Diego}
(\byear{2000})
\end{bbook}
\endbibitem

\bibitem{nair2010rectified}
\begin{bchapter}
\bauthor{\bsnm{Nair}, \binits{V.}},
\bauthor{\bsnm{Hinton}, \binits{G.E.}}:
\bctitle{Rectified linear units improve restricted boltzmann machines}.
In: \bbtitle{Icml}
(\byear{2010})
\end{bchapter}
\endbibitem

\bibitem{hendrycks2016gaussian}
\begin{botherref}
\oauthor{\bsnm{Hendrycks}, \binits{D.}},
\oauthor{\bsnm{Gimpel}, \binits{K.}}:
Gaussian error linear units (gelus).
arXiv preprint arXiv:1606.08415
(2016)
\end{botherref}
\endbibitem

\bibitem{kingma2014adam}
\begin{botherref}
\oauthor{\bsnm{Kingma}, \binits{D.P.}},
\oauthor{\bsnm{Ba}, \binits{J.}}:
Adam: A method for stochastic optimization.
arXiv preprint arXiv:1412.6980
(2014)
\end{botherref}
\endbibitem

\bibitem{li2020multipole}
\begin{barticle}
\bauthor{\bsnm{Li}, \binits{Z.}},
\bauthor{\bsnm{Kovachki}, \binits{N.}},
\bauthor{\bsnm{Azizzadenesheli}, \binits{K.}},
\bauthor{\bsnm{Liu}, \binits{B.}},
\bauthor{\bsnm{Stuart}, \binits{A.}},
\bauthor{\bsnm{Bhattacharya}, \binits{K.}},
\bauthor{\bsnm{Anandkumar}, \binits{A.}}:
\batitle{Multipole graph neural operator for parametric partial differential
  equations}.
\bjtitle{Advances in Neural Information Processing Systems}
\bvolume{33},
\bfpage{6755}--\blpage{6766}
(\byear{2020})
\end{barticle}
\endbibitem

\bibitem{ronneberger2015u}
\begin{bchapter}
\bauthor{\bsnm{Ronneberger}, \binits{O.}},
\bauthor{\bsnm{Fischer}, \binits{P.}},
\bauthor{\bsnm{Brox}, \binits{T.}}:
\bctitle{U-net: Convolutional networks for biomedical image segmentation}.
In: \bbtitle{International Conference on Medical Image Computing and
  Computer-assisted Intervention},
pp. \bfpage{234}--\blpage{241}
(\byear{2015}).
\bcomment{Springer}
\end{bchapter}
\endbibitem

\bibitem{wang2020towards}
\begin{bchapter}
\bauthor{\bsnm{Wang}, \binits{R.}},
\bauthor{\bsnm{Kashinath}, \binits{K.}},
\bauthor{\bsnm{Mustafa}, \binits{M.}},
\bauthor{\bsnm{Albert}, \binits{A.}},
\bauthor{\bsnm{Yu}, \binits{R.}}:
\bctitle{Towards physics-informed deep learning for turbulent flow prediction}.
In: \bbtitle{Proceedings of the 26th ACM SIGKDD International Conference on
  Knowledge Discovery \& Data Mining},
pp. \bfpage{1457}--\blpage{1466}
(\byear{2020})
\end{bchapter}
\endbibitem

\bibitem{driscoll2014chebfun}
\begin{botherref}
\oauthor{\bsnm{Driscoll}, \binits{T.A.}},
\oauthor{\bsnm{Hale}, \binits{N.}},
\oauthor{\bsnm{Trefethen}, \binits{L.N.}}:
Chebfun guide.
Pafnuty Publications, Oxford
(2014)
\end{botherref}
\endbibitem

\end{thebibliography}


\end{document}